%% file: BEV_main.tex
\documentclass[10pt,journal,compsoc]{IEEEtran}
\ifCLASSOPTIONcompsoc
  \usepackage[nocompress]{cite}
\else
  \usepackage{cite}
\fi

\ifCLASSINFOpdf
\else
\fi
\input{packages}
\input{macros}
\newcommand{\tabincell}[2]{\begin{tabular}{@{}#1@{}}#2\end{tabular}}
\usepackage{accents}
\newcommand{\vect}[1]{\accentset{\rightharpoonup}{#1}}

\usepackage{graphicx}

\begin{document}
%
\title{Vision-Centric BEV Perception: A Survey}

\author{Yuexin~Ma$^{\dag*}$, Tai~Wang$^\dag$, Xuyang~Bai$^\dag$, Huitong~Yang, Yuenan~Hou, \\Yaming~Wang, Yu~Qiao, Ruigang~Yang, Dinesh~Manocha, Xinge~Zhu$^*$

\IEEEcompsocitemizethanks{
\IEEEcompsocthanksitem Yuexin Ma is with ShanghaiTech University, Shanghai Engineering Research Center of Intelligent Vision and Imaging. E-mail: mayuexin@shanghaitech.edu.cn.
\IEEEcompsocthanksitem Xinge Zhu, Tai Wang are with the Chinese University of Hong Kong. E-mail: \{zx018, wt019\}@ie.cuhk.edu.hk.
\IEEEcompsocthanksitem Xuyang Bai is with Hong Kong University of Science and Technology. E-mail: xbaiad@connect.ust.hk.
\IEEEcompsocthanksitem Yuenan Hou is with Shanghai AI Lab. E-mail: \{houyuenan\}@pjlab.org.cn.
\IEEEcompsocthanksitem Yu Qiao is with Shenzhen Institute of Advanced Technology, Chinese Academy of Sciences, China, and also with Shanghai AI Lab, Shanghai, China. E-mail: yu.qiao@siat.ac.cn.
\IEEEcompsocthanksitem Huitong Yang is with ShanghaiTech University. E-mail: huitongy0126@gmail.com.
\IEEEcompsocthanksitem Ruigang Yang is with University of Kentucky. E-mail: ryang@cs.uky.edu
\IEEEcompsocthanksitem Dinesh Manocha, Yaming Wang are with the University of Maryland, College Park. E-mail: dmanocha@umd.edu, wym@umiacs.umd.edu
\IEEEcompsocthanksitem $\dag$ equal contributions and $^*$ corresponding authors
}
}

\markboth{IEEE TRANSACTIONS ON PATTERN ANALYSIS AND MACHINE INTELLIGENCE, VOL. X, NO. X, MMMMMMM YYYY}%
{}

%



\IEEEtitleabstractindextext{%
\input{abs}

}

\maketitle

\IEEEdisplaynontitleabstractindextext

%
\IEEEpeerreviewmaketitle

\input{intro}

\input{background}

\input{pv2bev}

\input{extension}
\input{conclusion}


%

%

%
%

\ifCLASSOPTIONcaptionsoff
  \newpage
\fi



%
\bibliographystyle{IEEEtran}
\bibliography{ref}

%
%

%

%
%
%
%




\end{document}

%% file: packages.tex
\usepackage{color,xcolor}
\usepackage{epsfig}
\usepackage{graphicx}

\usepackage[export]{adjustbox}
\usepackage{array}
\usepackage{booktabs}
\usepackage{colortbl}
\usepackage{float,wrapfig}
\usepackage{hhline}
\usepackage{multirow}
\usepackage{ragged2e}
\usepackage{subfigure}
\usepackage{caption}
\usepackage{comment}

\usepackage{amsmath,amsfonts,amsthm,amssymb}
\usepackage{bm}
\usepackage{nicefrac}
\usepackage{microtype}

\usepackage{changepage}
\usepackage{extramarks}
\usepackage{fancyhdr}
\usepackage{lastpage}
\usepackage{setspace}
\usepackage{soul}
\usepackage{xspace}
\usepackage{amsfonts}
\usepackage{adjustbox}
\usepackage{multirow}
\usepackage{pifont}
\usepackage{arydshln}
\usepackage{xcolor}
\usepackage{url}

\usepackage{algorithm, algorithmic}
\usepackage{enumerate}
\usepackage{todonotes}

\usepackage{bbm}


%% file: macros.tex
\newcolumntype{L}[1]{>{\raggedright\let\newline\\\arraybackslash\hspace{0pt}}m{#1}}
\newcolumntype{C}[1]{>{\centering\let\newline\\\arraybackslash\hspace{0pt}}m{#1}}
\newcolumntype{R}[1]{>{\raggedleft\let\newline\\\arraybackslash\hspace{0pt}}m{#1}}

\newcommand{\ignorethis}[1]{}

\makeatletter
\DeclareRobustCommand\onedot{\futurelet\@let@token\@onedot}
\def\@onedot{\ifx\@let@token.\else.\null\fi\xspace}

\def\eg{\emph{e.g}\onedot} 
\def\ie{\emph{i.e}\onedot}

\makeatother

\definecolor{MyDarkBlue}{rgb}{0,0.08,1}
\definecolor{MyDarkGreen}{rgb}{0.02,0.6,0.02}
\definecolor{MyDarkRed}{rgb}{0.8,0.02,0.02}
\definecolor{MyDarkOrange}{rgb}{0.40,0.2,0.02}
\definecolor{MyPurple}{RGB}{111,0,255}
\definecolor{MyRed}{rgb}{1.0,0.0,0.0}
\definecolor{MyGold}{rgb}{0.75,0.6,0.12}
\definecolor{MyDarkgray}{rgb}{0.66, 0.66, 0.66}



%% file: abs.tex
\begin{abstract}

In recent years, vision-centric Bird's Eye View (BEV) perception has garnered significant interest from both industry and academia due to its inherent advantages, such as providing an intuitive representation of the world and being conducive to data fusion. The rapid advancements in deep learning have led to the proposal of numerous methods for addressing vision-centric BEV perception challenges. However, there has been no recent survey encompassing this novel and burgeoning research field. To catalyze future research, this paper presents a comprehensive survey of the latest developments in vision-centric BEV perception and its extensions. It compiles and organizes up-to-date knowledge, offering a systematic review and summary of prevalent algorithms. Additionally, the paper provides in-depth analyses and comparative results on various BEV perception tasks, facilitating the evaluation of future works and sparking new research directions. Furthermore, the paper discusses and shares valuable empirical implementation details to aid in the advancement of related algorithms.

\end{abstract}

%% file: intro.tex
\IEEEraisesectionheading{\section{Introduction}\label{sec:introduction}}	 




\IEEEPARstart{A}ccurate and comprehensive understanding of surrounding environments, including dynamic objects and static infrastructure, is crucial for autonomous vehicles to make safe and effective driving decisions. Bird's eye view (BEV) 3D perception has attracted significant interest in recent years for two primary reasons. Firstly, BEV representations of the world, particularly in traffic scenarios, contain rich semantic information, precise localization, and absolute scales. These can be directly utilized by numerous downstream real-world applications, such as behavior prediction and motion planning. Secondly, BEV offers a physically interpretable approach for fusing information from different views, modalities, time series, and agents. As it represents the scene in a world coordinate system, multiple views of data captured by surrounding cameras can be integrated into a comprehensive BEV representation without additional stitching operations in overlapping areas. Concurrently, the temporal fusion of sequential visual data is also accurate and seamless, devoid of any distortion present existing in the perspective view. Moreover, other commonly employed acquisition sensors, such as LiDAR and radar, capture data in 3D space, which can be readily transformed to BEV and used for sensor fusion with cameras. It is worth mentioning that for the vehicle-vehicle and vehicle-infrastructure cooperative systems, BEV representation also plays a vital role in merging diverse information from multiple sources.


For cost-effective autonomous driving systems, vision-centric BEV perception remains long-standing challenges, as cameras are typically mounted on ego-vehicles parallel to the ground and facing outwards. Images are captured in a Perspective View (PV), which is orthogonal to BEV, and the transformation from PV to BEV is commonly referred to inverse perspective mapping. Over 30 years ago, the earliest work~\cite{IPM} attempted to tackle this problem by using a homography matrix to transform flat ground from PV to BEV directly in a geometric computing manner. These methods prevailed for years due to the computation efficiency until the rigid flat-world assumption no longer satisfy the requirements of autonomous driving in complex real-world scenarios, where 3D objects in the environment like vehicles possess height and consequently cause noticeable artifacts after transformation.

\begin{figure*}
    \centering
    \includegraphics[width=0.9\textwidth]{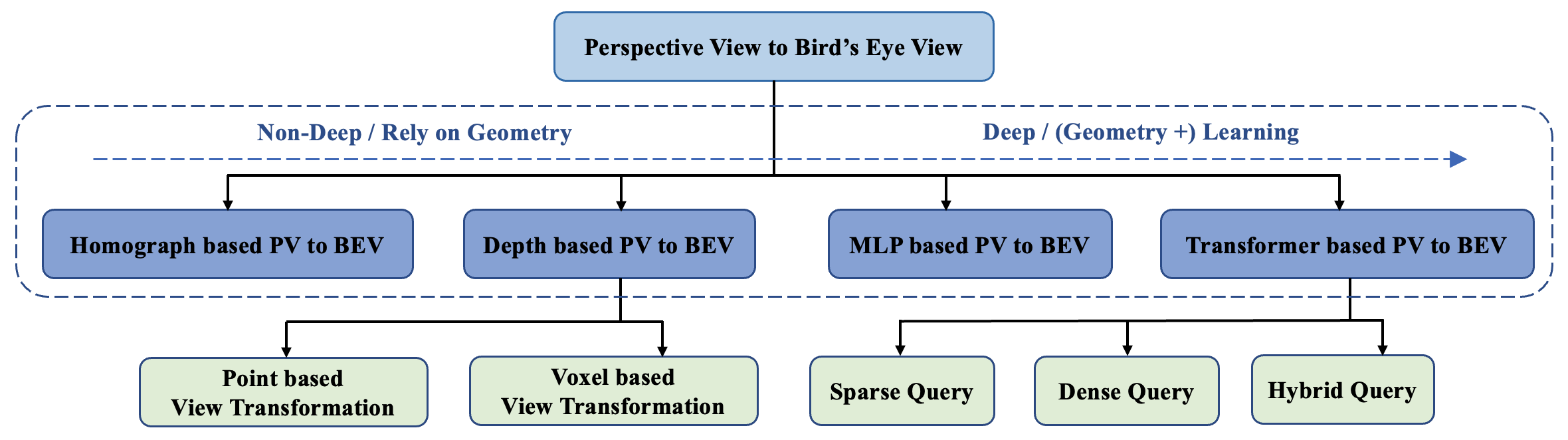}
    \vspace{-2ex}
    \caption{A taxonomy of algorithms for perspective view to bird's eye view. We categorize the methods for view transformation into four streams, following the development from non-deep approaches relying on geometry to deep ones involving learning. To clarify this development process and the differences among these streams, we write a separate sub-section for each stream to summarize the integration of subsequent methods with previous philosophies.}
    \vspace{-3ex}
    \label{fig:taxforbev}
\end{figure*}

With the advancements in data-driven methods in computer vision, numerous deep learning-based approaches have emerged in recent years to enhance vision-centric BEV perception by addressing the PV-BEV transformation challenge. These methods can be classified into three main streams based on their view transformation techniques: depth-based, MLP-based, and transformer-based approaches. For depth-based methods, lifting 2D features to 3D space via explicit or implicit depth estimation serves as a major solution. For each pixel in the image, a ray emanates from the camera and intersects objects in the real world. Rather than directly mapping pixels to BEV, an alternative approach calculates the depth distribution for each pixel, elevates the 2D features to 3D with this distribution, and subsequently obtains the BEV representations from 3D through dimensionality reduction. Various assumptions are made regarding depth, such as an exact value, a uniform distribution along the ray, or a categorical distribution along the ray. And the depth supervision stems from either explicit depth values or task supervision at the end. Considering that deep neural networks have made significant strides in addressing computer vision tasks by acting as complex mapping functions that transform input to output with different modalities, dimensions, representations, etc., one straightforward approach is to utilize a variational encoder-decoder or MLP to project PV features to BEV. MLP-based methods are simple to implement but struggle to generalize in complex scenarios with occlusions and multi-view input settings. Actually, the aforementioned methods adopt a bottom-up strategy, handling the transformation in a forward manner. Another prominent type of method, transformer-based approaches, employ a top-down manner, directly constructing BEV queries and searching for corresponding features in perspective images using the cross-attention mechanism. Sparse, dense, or hybrid queries are proposed to accommodate various downstream tasks. These transformer-based methods possess a strong relation modeling ability and data-dependent properties, achieving impressive performance.

As the Fig.~\ref{fig:taxforbev} shows, vision-centric BEV perception has evolved from non-deep approaches relying on geometry to deep ones involving learning. Given the massive remarkable achievements in both academia and industry related to this area, we present a comprehensive review of recent progress to facilitate further research. The primary contributions of this work can be summarized as follows:







\begin{enumerate}

\item To the best of our knowledge, this is the first survey to review recent progress in addressing the view transformation between the perspective view and bird's eye view.

\item We present the most up-to-date methods of vision-centric BEV perception, categorizing them based on their core view transformation strategy and downstream vision tasks. And we also provide detailed analysis and comparison of the performance and limitations of these methods.

\item We propose additional extensions of BEV perception work, including multi-task learning strategies, fusion operations in BEV, semantic occupancy prediction, and practical training techniques, to support the implementation and development of related approaches.

\end{enumerate}

This paper is organized as follows. Section~\ref{background} introduces the background of vision-centric BEV perception.
Section~\ref{sec::homograph},\ref{sec::depth},\ref{sec:MLP},\ref{sec:transformer-based} surveys homograph-based, depth-based, MLP-based, and transformer-based methods, respectively, including the introduction, analysis, and comparison of popular methods and the summary of pros and cons. Specifically, we also conclude the combination ways with other streams of methods for each category. 
Section~\ref{extension} presents rich extensions under BEV. Section~\ref{conclusion} concludes this paper. We provide a regularly updated summary page at {https://github.com/4DVLab/Vision-Centric-BEV-Perception}.

%% file: background.tex
\section{Background}
\label{background}

We discuss four aspects of the background to this problem, including the task definition and conventional solutions for different tasks, frequently used datasets, common evaluation metrics, and dataset-specific metrics.

\begin{table*}[!t]
\centering
\caption{Detailed information of benchmarks that can be used for BEV-based 3D detection. Some datasets can be used for multiple tasks, and here we report the number of samples for 3D detection ({\it e.g.}, there are more than 40K images in KITTI, but only 15K of them are used for 3D detection). $^{*}$: Data in brackets denote the number of classes used in the official benchmarks.}
\vspace{-2ex}
\resizebox{0.9\linewidth}{!}{
\begin{tabular}{rcrrrrcccccccc}
\toprule
Dataset & \# Views &  Train &  Val &  Test & \# Boxes & \# Scenes & \# Classes$^*$ &  Night/Rain & Stereo & Temporal & LiDAR & Benchmark\\
\midrule
KITTI~\cite{kitti} & 1 & 7,418 & - & 7,518 & 200K & - & 8(3)  & $\times$/$\times$ & $\surd$ & $\surd$ & $\surd$ & $\surd$ \\
Argoverse~\cite{argoverse} & 
7 & 39,384 & 15,062 & 12,507 & 993K & 113 & 15 & $\surd$/$\surd$  & $\surd$ & $\surd$ & $\surd$ & $\surd$\\
Lyft L5~\cite{lyft}& 6 & 22,690 & - & 27,468 & 1.3M & 366 & 9 & $\times$/$\times$  & $\times$ & $\surd$ & $\surd$ & $\times$\\
H3D~\cite{h3d}& 3 &  8,873 & 5,170 & 13,678 & 1.1M & 160 & 8 &  $\times$/$\times$  & $\times$ & $\surd$ & $\surd$ & $\times$\\
nuScenes~\cite{nuscenes} & 6 &  28,130 & 6,019 & 6,008 & 1.4M & 1,000 & 23(10) & $\surd$/$\surd$  & $\times$ & $\surd$ & $\surd$ & $\surd$\\
Waymo Open Dataset~\cite{waymo} & 5 & 122,200 & 30,407 & 40,077 & 12M & 1,150 & 4(3) & $\surd$/$\surd$  & $\times$ & $\surd$ & $\surd$ & $\surd$\\
CityScapes 3D~\cite{cityscapes3d}& 
1 & 2,975 & 500 & 1,525 & 40K &  - & 8(6) & $\surd$/$\surd$ & $\surd$ & $\times$ & $\times$ & $\surd$\\
\bottomrule
\end{tabular}}
\label{dataset_summary}
\vspace{-2ex}
\end{table*}

\subsection{Task Definition of Vision-Centric BEV Perception}


\emph{Vision-centric BEV perception} refers to the concept that, given an input image sequence $\mathbf{I} \in \mathbb{R}^{N \times V \times H \times W \times 3}$, algorithms need to transform these perspective-view inputs into BEV features and perform perception tasks such as detecting 3D bounding boxes of objects or generating semantic maps of the surrounding environment in the bird's eye view. Here, $N, V, H, W$ represent the number of frames, views, height, and width of the input image, respectively.


\subsection{Datasets \& Common Evaluation Metrics}
KITTI~\cite{kitti}, nuScenes~\cite{nuscenes}, and Waymo Open Dataset (WOD)~\cite{waymo} are the three most influential benchmarks for BEV-based 3D perception. KITTI is a renowned benchmark for 3D perception, consisting of 3712, 3769, and 7518 samples for training, validation, and testing, respectively. It provides both 2D and 3D annotations for cars, pedestrians, and cyclists. Detection is divided into three levels, i.e., easy, moderate, and hard, based on the size of detected objects, occlusion, and truncation levels.
NuScenes contains 1000 scenes, each with a duration of 20 seconds. Each frame includes six calibrated images covering a 360-degree horizontal field of view (FOV), making nuScenes one of the most widely used datasets for vision-based BEV perception algorithms. WOD is a large-scale autonomous driving dataset with 798 sequences, 202 sequences, and 150 sequences for training, validation, and testing, respectively. In addition to the aforementioned three datasets, other benchmarks such as Argoverse, H3D, and Lyft L5 can also be utilized for BEV-based perception. Detailed information is summarized in Table~\ref{dataset_summary}.

As to the common evaluation metrics, the most commonly used criterion for BEV Detection is average precision (AP) and the mean average precision (mAP) over different classes or difficulty levels. For BEV Segmentation, IoU for each class and mIoU over all classes are frequently used as the metrics.




\subsection{Dataset-Specific Metrics}

\noindent \textbf{KITTI.}\quad KITTI makes several modifications to the AP metric. First, the IoU is calculated in the 3D space. Second, it includes 40 recall positions instead of 11 and removes the recall position at 0. Specifically, the used $R_{40}$ is $\{1/40, 2/40, ..., 1\}$. In addition, since the height of objects is not very important in the BEV, it also introduces BEV AP, and the IoU is calculated on the ground plane instead of in 3D space.
Moreover, KITTI also introduces a new metric, \ie, Average Orientation Similarity (AOS), which evaluates the quality of orientation estimation. The definition of AOS is provided as: $\mathbf{AOS} = \frac{1}{|R|}\sum_{r \in R}{\max_{r':r'\geq r}c(r')}$. The orientation similarity $c(r)$ is the normalized variant of the cosine similarity, the definition of which is: $\mathbf{c(r)} = \frac{1}{|B(r)|}\sum_{i \in |B(r)|}{\frac{1+\cos{\Delta \theta_{i}}}{2} \delta_{i}}$, where $B(r)$ is the set of all detection results at recall $r$, $\Delta \theta_{i}$ is the difference of orientation prediction and ground-truth orientation of detection and $\delta_{i}$ is the penalty term to penalize duplicate predictions on the same object. It is noteworthy that all AP metrics are calculated independently for each difficulty level and each class.

\noindent \textbf{NuScenes.}\quad In contrast to conventional AP calculation, which uses IoU to select TP, nuScenes leverages the 2D center distance on the ground plane to match the predictions and ground truths with a certain distance threshold $d$, \eg, 2 meters. In addition, nuScenes calculates AP as the normalized area under the precision-recall curve for recall and precision over 10\%. Finally, the mAP is calculated over all matching thresholds, $\mathbb{D}=\{0.5,1,2,4\}$ meters and all classes $\mathbb{C}$: ${\rm mAP} = \frac{1}{|\mathbb{C}||\mathbb{D}|}\sum_{c \in \mathbb{C}}\sum_{d\in \mathbb{D}}{\rm AP}_{c, d}$.

However, this metric only considers the 3D position of objects and ignores the effects of both dimension and orientation. To compensate for it, nuScenes also proposes several True Positive metrics (TP metrics) that aim to measure each prediction separately using all true positives (determined under the center distance $d = 2m$ during matching). These metrics are Average Translation Error, Average Scale Error, Average Orientation Error, Average Velocity Error and Average Attribute Error.
For each TP metric, nuScenes also computes the mean TP metric (mTP) over all categories: ${\rm mTP}_k = \frac{1}{|\mathbb{C}|} \sum_{c \in \mathbb{C}} \mathrm{TP}_{k, c}$,
where $\mathrm{TP}_{k, c}$ denotes the $k^{th}$ TP metric for class $c$. nuScenes further proposes the nuScenes Detection Score (NDS), which is the combination of the mAP and the mTP${_k}$ metrics: ${\rm NDS} = \frac{1}{10} [5\cdot {\rm mAP}+\sum_{k=1}^5(1-\min(1, {\rm mTP}_{k}))]$.

\begin{figure*}
    \centering
    \includegraphics[width=1\textwidth]{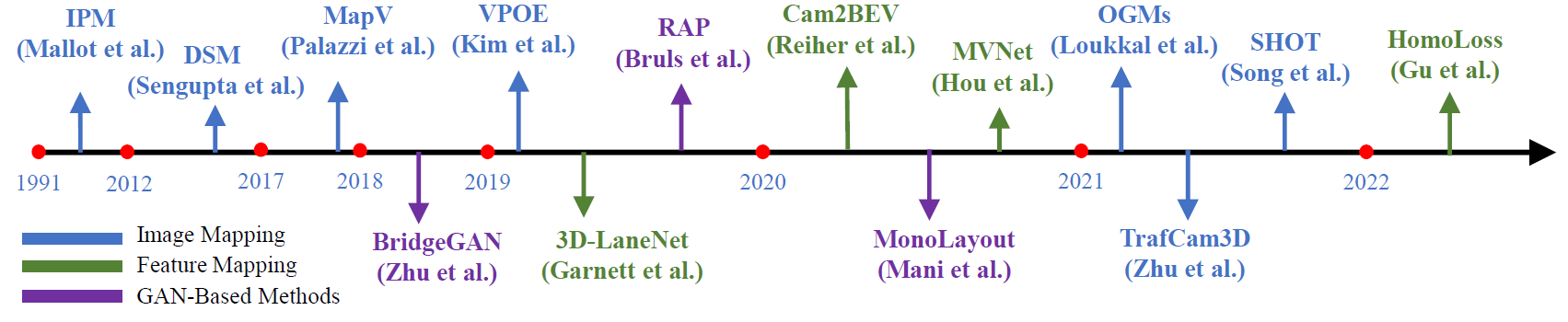}
    \vspace{-4ex}
    \caption{Chronological overview of homograph based PV to BEV methods.}
    \vspace{-2ex}
    \label{fig:homograph-overview}
\end{figure*}

\noindent \textbf{Waymo Open Dataset.}\quad Waymo Open Dataset replaces $\mathbb{R}_{11}$ with $\mathbb{R}_{21}=\{0, 1/20, 2/20, 3/20, ..., 1\}$ when calculating the AP metric. In addition, it incorporates the heading prediction into the AP metric and puts forward the Average Precision weighted by Heading (APH) as its main metric. Specifically, APH takes the heading information into account when calculating the precision. Each TP is weighted by the heading accuracy, the definition of which is given as $\min(|\theta-\hat{\theta}|, 2\pi-|\theta-\hat{\theta}|)/\pi$, where $\theta$ and $\hat{\theta}$ are the predicted heading angle and the ground-truth label whose range is $[-\pi,\pi]$. APH jointly assesses the performance of both 3D object detection and orientation estimation, while AOS only evaluates the quality of orientation estimation.

Recently, the Waymo team has proposed a new metric~\cite{let-3d-ap} for image-based 3D detection, \ie, Longitudinal Error Tolerant 3D Average Precision (LET-3D-AP), which rewards predictions with small lateral errors but relatively large longitudinal errors. The intuition is that these predictions are affected by depth estimation errors but still provide valuable information about the environment. In other words, this metric is designed to be more tolerant with respect to depth estimation errors. Given a ground truth bounding box with center $\vect{G}=[x_{g},y_{g},z_{g}]$ and a predicted box with center $\vect{P}=[x_{p},y_{p},z_{p}]$, they first define the longitudinal affinity $a_l(\vect{P}, \vect{G})$, which is the score for matching predicted bounding boxes with ground truth bounding boxes given a tolerance for the longitudinal error: $a_l(\vect{P}, \vect{G}) =1-\min{\left(|\vect{e_{\text{lon}}}(\vect{P}, \vect{G})|/T_{\text{l}}, 1.0\right)}$,
where $T_{l}=\text{max}(T_l^p  \times |G|, T_l^m)$, $T_l^p$ is the longitudinal tolerance percentage and $T_l^m$ controls the tolerance for near range objects.
They also propose the Longitudinal Error Tolerant Intersection-over-Union (LET-IoU), which is calculated by compensating for the longitudinal error. Specifically, they first project the ground truth center onto the line of sight from the sensor to the prediction: $\vect{P_{\text{aligned}}}=(\vect{G}\cdot\vect{u_P}) \times \vect{u_P}$, 
where $\vect{u_P}=\vect{P}/|\vect{P}|$ is the unit vector along the line of sight to prediction center. Then, the LET-IOU is computed via: $\text{LET-IoU}(P, G) = \text{3D-IoU}(P_{\text{aligned}}, G)$,
where $P_{\text{aligned}}$ is the predicted bounding box with aligned center~$\vect{P_{\text{aligned}}}$.
After that, they perform the bipartite matching by taking the longitudinal error tolerance into account. The bipartite matching weight $W(i, j)$ is set as $a_l(P(i), G(j)) \times \text{LET-IoU}(P(i), G(j))$ if $a_l > 0$ and \text{LET-IoU} is larger than the pre-defined IoU threshold $T_\text{iou}$. Otherwise it is set as zero.
After the bipartite matching, TP, FP and FN are determined, which can be used to compute the precision and recall. The LET-3D-AP (Average Precision with Longitudinal Error Tolerance) can be computed by: $\text{LET-3D-AP}=\int_0^1 p(r)dr$,
where $p(r)$ is the precision value at recall $r$. 
In their paper, they also introduce the LET-3D-APL (Longitudinal Affinity Weighted LET-3D-AP) and this metric penalizes the predictions that do not overlap with any ground truth. Please refer to the original paper~\cite{let-3d-ap} for more details.

%% file: pv2bev.tex

\input{IPM}
\label{IPM}
\input{Depth}

\label{depth}


\input{MLP}
\label{MLP}
\input{Transformer}

\label{transformer}


%% file: IPM.tex
\section{Homograph based PV to BEV}
\label{sec::homograph}

A traditional and intuitive method for transforming PV to BEV is utilizing the inherent geometric projection relationship between two views. Inverse Perspective Mapping (IPM)~\cite{IPM} was proposed to address this challenging mapping problem, with the additional constraint that inversely mapped points lie on a horizontal plane.
\subsection{Basic Usage in Different Stages}
IPM is the pioneering work in warping a front-view image to a top-view image, thus intuitively being exploited in the preprocessing or post-processing at first. The transformation involves applying a camera rotation homography followed by anisotropic scaling~\cite{hartley2003multiple}. The homography matrix can be derived from the camera's intrinsic and extrinsic parameters. Some methods~\cite{GeoAppro} employ Convolutional Neural Networks (CNNs) to extract semantic features from the perspective-view image and estimate the vertical vanishing points and ground plane vanishing lines (horizon) in the image to determine the homography matrix. After the IPM operation, numerous downstream perception tasks, such as optical flow estimation, detection, segmentation, motion prediction, and planning, can be performed based on the BEV image. VPOE~\cite{VPOE} integrates Yolov~\cite{yolov3} as the detection backbone to estimate vehicle position and orientation in BEV. Using a synthetic dataset, ~\cite{Palazz} maps detections from a dashboard camera view onto a BEV occupancy map of the scene by IPM as well. In practical applications, the camera's intrinsic and extrinsic parameters may be unknown, and TrafCam3D~\cite{TrafCam3D} proposes a robust homography map based on a dual-view network architecture to mitigate IPM distortion.

Instead of applying IPM in preprocessing or postprocessing, some approaches opt to use it to transform feature maps during network training. Cam2BEV~\cite{cam2BEV} obtains the holistic BEV semantic map by applying IPM to transform the feature map of images captured by multiple vehicle-mounted cameras. MVNet~\cite{MVNet} projects 2D features into the shared BEV space based on IPM to aggregate multi-view features and employs large convolution kernels to address occlusion issues in pedestrian detection. Focusing on predicting the 3D layout of lanes from a single image, 3D-LaneNet~\cite{3D-LaneNet} does not assume camera height and trains an additional network branch in a supervised manner to estimate the homography matrix. It then adopts projective transformation on different scales of feature maps. Gu et al.~\cite{Homography-Loss} apply 2D detection predictions to globally optimize 3D boxes, and a Homography loss is proposed to embed geometric constraints between 2D and BEV space.

\subsection{Limitations and Solutions}

Since IPM heavily relies on the flat-ground assumption, IPM-based approaches typically struggle to accurately detect objects situated above the ground plane, such as buildings and vehicles. Some methods utilize semantic information to reduce distortions. OGMs~\cite{OGM} transforms the footprint segmentation results of vehicles in PV to BEV to adhere to the flat ground hypothesis implied by the homography, thus avoiding distortion caused by the vehicle body being located above the ground. Building on this concept, BEVStitch~\cite{BEVstitch} uses two branches to segment footprints of vehicles and roads, transforming them to BEV using IPM, respectively, and then stitches them on BEV to construct a complete road map. DSM~\cite{DSM} performs image semantic segmentation in the perspective view first and then uses homography to construct the semantic map in BEV. In particularly, SHOT~\cite{SHOT} processes pedestrians by projecting different parts of pedestrians onto various ground levels using multiple homography matrices.

Owing to the significant gap and severe deformation between the frontal view and bird's-eye view, relying solely on IPM is inadequate for generating distortion-free images or semantic maps in BEV. Generative Adversarial Network (GAN)~\cite{GAN} is exploited to enhance the authenticity of the generated BEV features or images. BridgeGAN~\cite{BridgeGAN} takes the homography view as an intermediate view and proposes a multi-GAN based model to learn the cross-view translation between PV and BEV.  
The subsequent work~\cite{2D3DLifting} addresses the monocular 3D detection problem by conducting 2D detection on BEV and aligning the results with the ground plane estimation to produce the final 3D detections. MonoLayout~\cite{Monolayout} also employs GAN to generate the information about invisible places and estimates the scene layout with dynamic objects. RAP~\cite{RAP} introduces an incremental GAN to learn more reliable IPM for a front-facing camera using robust real-world labels, which significantly alleviates the stretching of distant objects.


\subsection{Summary}

Homography-based methods (Fig.~\ref{fig:homograph-overview}) rely on the physical mapping of flat ground between perspective view and bird's-eye view, offering good interpretability. IPM serves as a tool for image projection or feature projection for downstream perception tasks. To minimize distortion in areas above the ground plane, semantic information is thoroughly utilized, and GAN is widely employed to enhance the quality of BEV features. The core mapping procedure, which involves straightforward matrix multiplication, does not require learning and is an efficient choice. However, IPM only addresses part of the PV-BEV transformation problem through a rigid flat-ground assumption, limiting its application in real 3D scenarios. A comprehensive and effective BEV mapping for the entire content of PV remains to be achieved.

%% file: Depth.tex
\begin{figure*}
    \centering
    \includegraphics[width=1.0\textwidth]{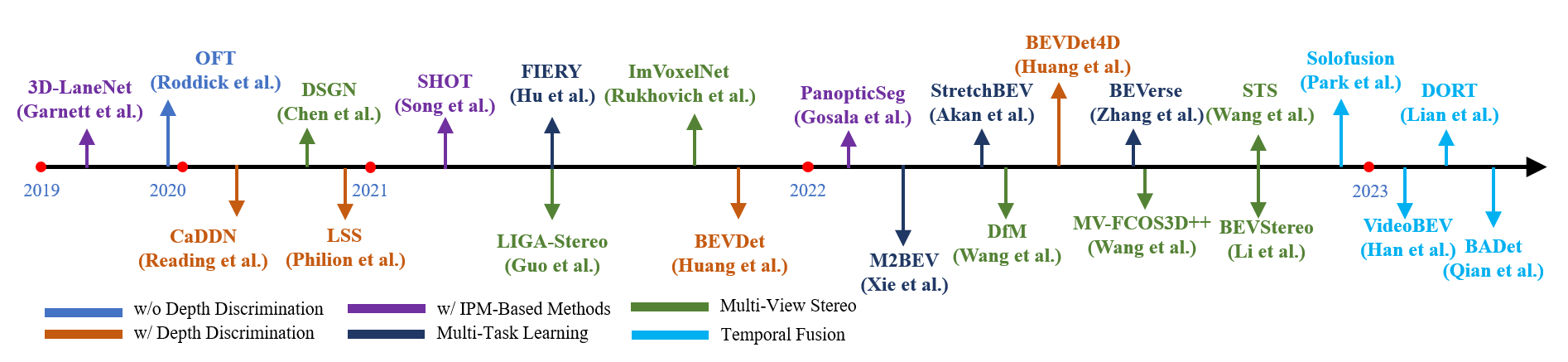}
    \vspace{-4ex}
    \caption{Chronological overview of depth based PV to BEV methods.}
    \vspace{-2ex}
    \label{fig:depth-based-overview}
\end{figure*}

\begin{figure}
	\centering
	    \subfigure[Pseudo-LiDAR~\cite{pseudolidar} pipline]
	    {\centering
	     \includegraphics[width=0.92\columnwidth]{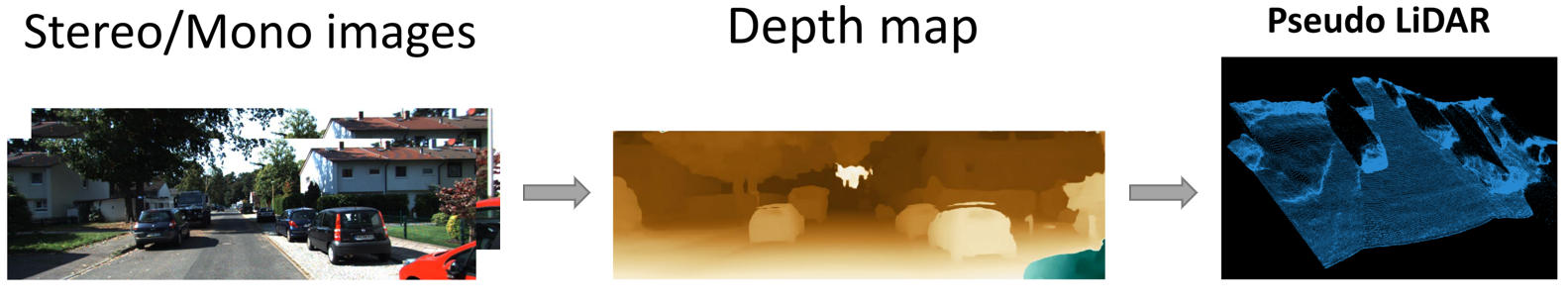}} 
	    \subfigure[Pseudo-LiDAR++~\cite{pseudolidar++} for more accurate depth estimation]
	    {\centering
	     \includegraphics[width=0.92\columnwidth]{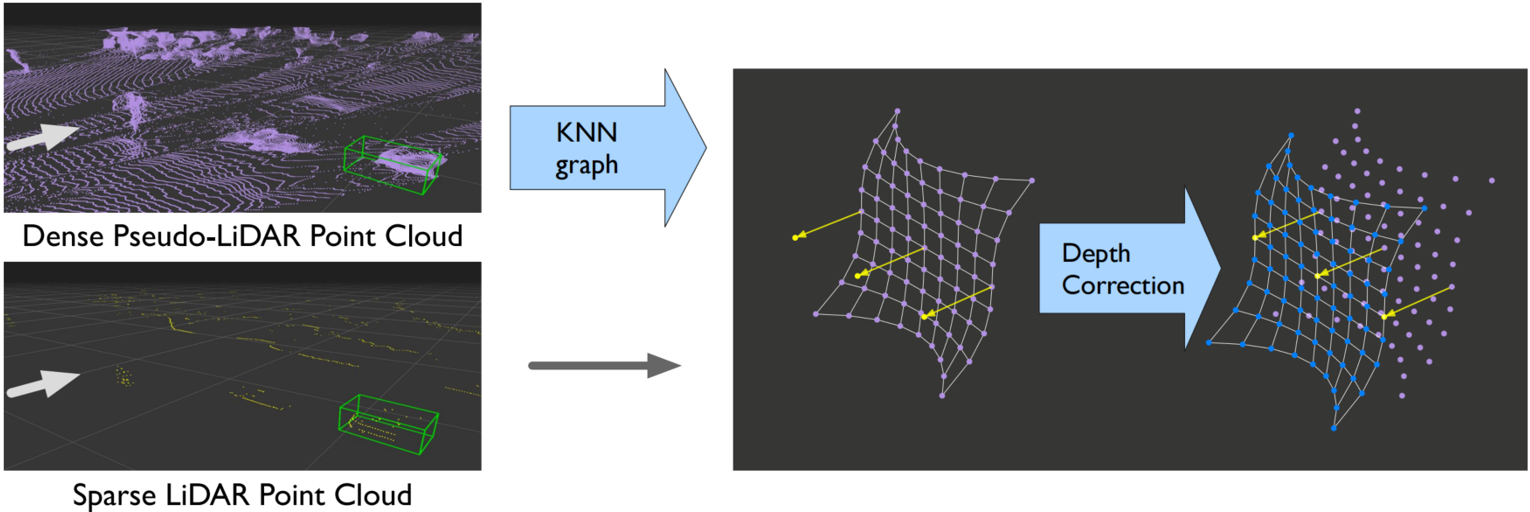}} 
      \vspace{-2ex}
	     \caption{Point-based methods transform 2D image pixels to Pseudo-LiDAR and use LiDAR-based approaches for 3D object detection.}
	     \vspace{-2ex}
	\label{fig:pseudo-lidar}
\end{figure}

\vspace{-2ex}
\begin{figure}[hb]
	\centering
	    \subfigure[OFT~\cite{OFT} does not predict the depth distribution and scatters the same image feature along a ray.]
	    {\centering
	     \includegraphics[width=0.85\columnwidth]{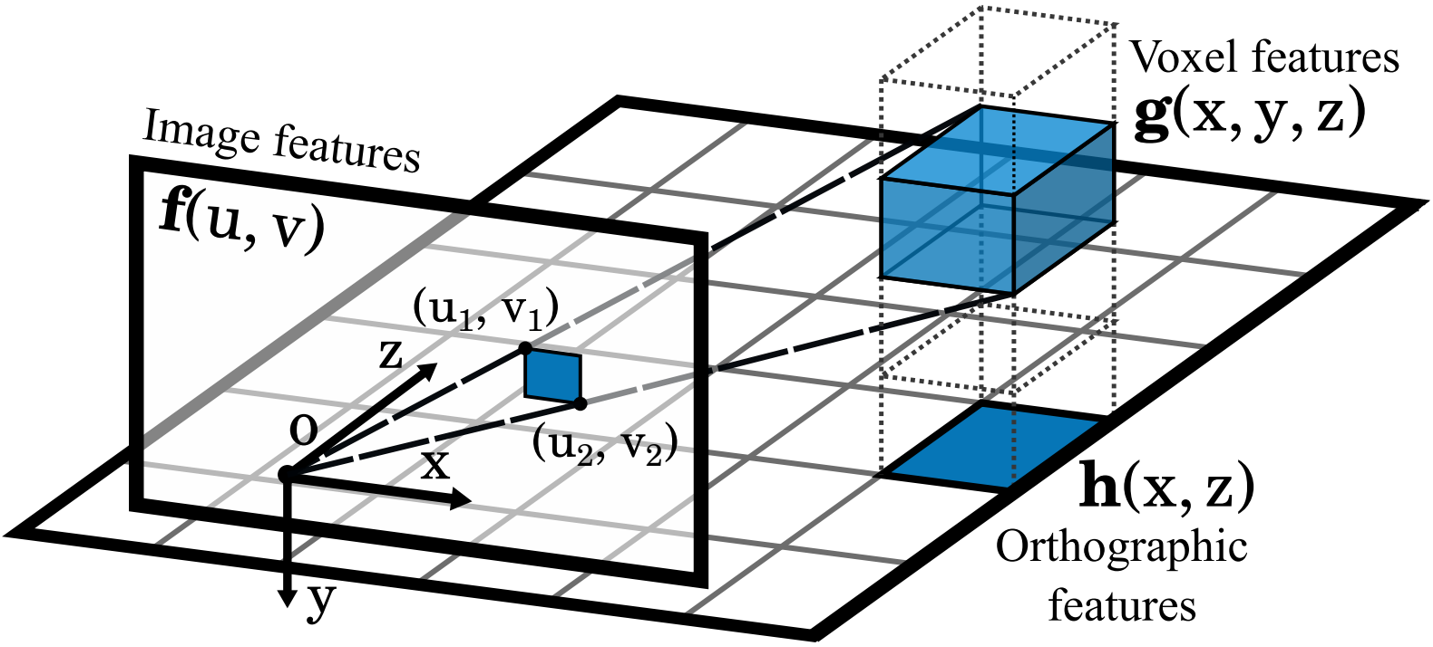} }\\
      	    \subfigure["lift" step in "Lift,Splat,Shoot"]
	    {\centering
	     \includegraphics[width=0.85\columnwidth]{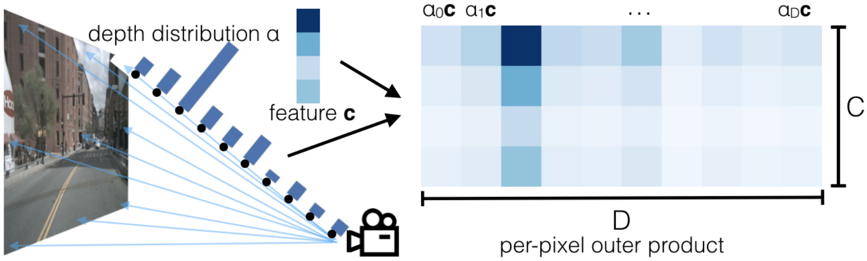}}
	 \vspace{-2ex}    
      \caption{The comparison of depth distribution between LSS~\cite{LSS} and OFT~\cite{OFT}. }
	     \vspace{-2ex}
	\label{fig:explicit-depth}
\end{figure}

\section{Depth based PV to BEV}
\label{sec::depth}

IPM-based methods are constructed on the assumption that all points lie on the ground plane. While this offers a feasible approach for bridging the 2D perspective space and the bird's-eye view of 3D space, it sacrifices crucial height differentiation. To address this limitation, depth information is required to elevate 2D pixels or features into 3D space. Driven by this insight, a significant trend after IPM-based methods for PV-BEV transformation is founded on depth predictions. In the following discussion, we will first compare the technical designs in these approaches, including the method of view transformation, the inclusion of depth supervision, and the integration with IPM-based methods. Lastly, we will explore the inherent advantages of such methods in multi-view scenarios.

\subsection{Point-Based View Transformation}

Depth-based PV-to-BEV methods are inherently built on an explicit 3D representation (Fig.~\ref{fig:depth-based-overview}). Similar to LiDAR-based 3D perception~\cite{zhu2021cylindrical,CenterPoint,qi2017pointnet++,cong2022stcrowd}, these methods can be classified into two categories based on the representation used: point-based and voxel-based methods.
Point-based methods directly utilize depth estimation to convert pixels into point clouds, scattered in continuous 3D space. These methods are more straightforward and can easily integrate mature techniques from monocular depth estimation and LiDAR-based 3D detection. Pioneering work, Pseudo-LiDAR~\cite{pseudolidar} (Fig.\ref{fig:pseudo-lidar}), initially converts depth maps into pseudo-LiDAR points, which are then fed into state-of-the-art LiDAR-based 3D detectors. Another groundbreaking work, Pseudo-LiDAR++\cite{pseudolidar++}, improves depth accuracy using a stereo depth estimation network and a loss function. AM3D~\cite{AM3D} suggests augmenting pseudo-point clouds with complementary RGB features. PatchNet~\cite{patchnet} examines the differences between depth maps and 3D coordinates, proposing the integration of 3D coordinates as additional input data channels to achieve comparable results.
However, such methods share two common issues: 1) Data leakage problem. Mistakenly involving the data from KITTI Depth Estimation Benchmark for depth estimator pretraining leads to data leakage to downstream 3D detection evaluation, resulting in incorrect high performance of such methods on the validation set, as analyzed in \cite{pseudolidar,arewemiss3dconf}. 2) Due to the gradient cut-off between the pseudo-LiDAR generation stage and the subsequent 3D detection stage, this pipeline is complex for both training and deployment, meanwhile bounded by the performance of depth estimation. 3) The generated pseudo-LiDAR is typically inaccurate and thus misleading. It is also denser than real LiDAR points, bringing a large computation burden for the 3D detection stage. E2E Pseudo-LiDAR~\cite{e2e-pseudolidar} introduces a Change-of-Representation (CoR) module to address the second problem, but follow-up works~\cite{CADDN,DD3D} further prove their inferiority to voxel-based methods in these aspects.


\subsection{Voxel-Based View Transformation}

In comparison to point clouds, which are distributed in continuous 3D space, voxels provide a more efficient representation for 3D scene understanding. They discretize the 3D space to construct a regular structure for feature transformation, allowing subsequent BEV-based modules to be directly appended. Although sacrificing local spatial precision, voxels have proven to be more effective at covering large-scale scene structure information and are compatible with end-to-end learning paradigms for view transformation. 

Specifically, this scheme typically scatters 2D \emph{features} (rather than points) at the corresponding 3D locations directly with depth guidance. Prior works achieve this by taking the outer product of the 2D feature map with a corresponding predicted depth distribution. Early works assume that the distribution is uniform, \emph{i.e.}, all the features along a ray are the same, as in OFT~\cite{OFT} (Fig.~\ref{fig:explicit-depth}). This early work constructs an internal representation to determine which image features are relevant to the location on the orthographic bird's eye view map. It creates a 3D voxel feature map, defined over a uniformly spaced 3D lattice, and fills the voxel by accumulating features over the area of the projected corresponding image feature map. The orthographic feature map is then obtained by summing voxel features along the vertical axis, and a deep convolutional neural network extracts BEV features for 3D object detection. It is worth noting that for each pixel on the image, the network predicts the same representation for each point in 3D assigned to it, \ie, predicting a uniform distribution over depth. This category of methods typically does not require depth supervision and can learn depth or 3D location information in the networks after view transformation in an end-to-end manner.

In contrast, another paradigm explicitly predicts the depth distribution and uses it to attentively construct the 3D feature. LSS~\cite{LSS}, as shown in Fig.~\ref{fig:explicit-depth}, is representative of this approach. It predicts a categorical distribution over depth and a context vector, and their outer product determines the feature at each point along the perspective ray, more accurately approximating the real depth distribution. Furthermore, it fuses predictions from all cameras into a single cohesive representation of the scene, which is more robust to calibration errors. BEVDet~\cite{Bevdet} follows LSS paradigm and proposes a framework for multi-view camera-only 3D detection on BEV, consisting of an image-view encoder, a view transformer, a BEV encoder, and a detection head. A new version, BEVDet4D~\cite{Bevdet4d}, exploits temporal cues in multi-camera-based 3D detection. Specifically, this method retains the intermediate BEV feature of the previous frame and concatenates it with the ones generated by the current frame.

\subsection{Depth Supervision}
Previous studies have shown that, when using predicted depth distributions to lift 2D features, the accuracy of this distribution is important. CaDDN~\cite{CADDN} leverages a classical method to interpolate sparse depth maps derived from projected LiDAR points and exploits them to supervise the prediction of depth distribution. It shows that this supervision and a loss function that encourages sharp distribution prediction are critical in this type of method. Other methods for binocular-based 3D detection, DSGN~\cite{dsgn}, and LIGA-Stereo~\cite{liga}, also rely on similar supervisions, where \emph{sparse} LiDAR depth maps are more effective. Other works that do not utilize depth labels can only learn such 3D localization or depth information from sparse instance an-notations, which is much more difficult for network learning. 
Apart from incorporating depth supervision in the detection framework, DD3D~\cite{DD3D} and MV-FCOS3D++~\cite{wang2022mvfcos3d++} point out that the pretraining of depth estimation and monocular 3D detection can significantly enhance the representation learning of a 2D backbone. Many previously mentioned BEV-based methods~\cite{Bevdet,Bevdet4d} also benefit from these pretraining backbones. More details will be presented in Sec.~\ref{implementation}.

\subsection{Multi-View Aggregation for Stereo Matching}
In addition to monocular depth estimation, stereo matching can predict more accurate depth information in camera-only perception. It relies on the baseline naturally formed by suitable multi-view settings. Among them, the binocular setting is the most common and well-studied one, and it features large overlap regions and only a small horizontal offset for establishing the suitable multi-view setting.
For comparison, in a general multi-view setting used in previous works~\cite{LSS,Bevdet}, \emph{e.g.}, surround-view cameras are mounted on an autonomous vehicle, the overlap regions across adjacent views are usually very small because the main target is to cover the entire space with fewer cameras. In this case, depth estimation relies primarily on monocular understanding, and BEV-based methods are only superior in terms of the simplicity and unification for multi-view perception.

In contrast, they have more important merits for depth estimation in binocular cases. Recent binocular methods, such as DSGN~\cite{dsgn} and LIGA-Stereo~\cite{liga}, typically use a plane-sweep representation for stereo matching and depth estimation. Then they sample the voxel and BEV feature from the plane-sweep feature volume and perform 3D detection thereon. Other methods targeting multi-view settings, such as ImVoxelNet~\cite{imvoxelnet}, also shows the effectiveness of such voxel-based formulation in indoor scenes, where the overlapped regions are also larger across adjacent regions. In addition, for consecutive frames, two temporally adjacent images can also satisfy such conditions. DfM~\cite{wang2022dfm} analyzes this problem theoretically and adopts similar methods to achieve more accurate monocular 3D detection from videos. Recent works~\cite{wang2022sts,li2022bevstereo,solofusion,videobev,dort,badet} further explore better practices, including how to leverage different frames during training and inference and how to model object motion in temporal multi-view stereo, along this direction under the context of multi-view 3D perception.

\vspace{-2ex}

\subsection{Combination with Previous Streams}

As previously discussed, IPM-based methods perform well and efficiently in flat-ground scenarios, requiring only a few parameters to learn. Methods that do not rely on explicit depth prediction and supervision are suitable for feature aggregation along the vertical direction. PanopticSeg~\cite{panopticseg} takes advantage of both strengths and proposes a dense transformer module for panoptic segmentation. This module is composed of a flat transformer that uses IPM, followed by error correction to generate the flat BEV features, and a vertical transformer that employs a volumetric lattice to model the intermediate 3D space, which is then flattened to produce the vertical BEV features.

\vspace{-3ex}
\subsection{Summary}
Depth-based view transformation methods are usually built on an explicit 3D representation, quantized voxels, or point clouds scattering in continuous 3D space. Voxel-based methods use a uniform depth vector or the explicitly predicted depth distribution to lift 2D features to a 3D voxel space and perform BEV-based perception thereon. In contrast, point-based methods convert the depth prediction to a pseudo-LiDAR representation and then use custom networks for 3D detection. Table~\ref{tab:3ddet-kitti},~\ref{tab:3ddet-nus} present results achieved by this type of method. We can observe that:
\begin{itemize}
    \item Earlier methods usually exploit pseudo-LiDAR rep-resentation for straightforward usage of 3D detectors in the second stage; however, they suffering from the model complexity and lower performance caused by the difficulty of generalizable end-to-end training.
    \item Recent methods pay more attention to voxel-based methods due to their computation efficiency and flexi-bility. This representation has been widely adopted in camera-only methods for different tasks.
    \item Depth supervision is important to such depth-based methods because accurate depth distribution can provide essential cues when converting perspective-view features to bird’s eye view.
    \item Exploring the potential benefits of temporal modeling is a promising direction, as analyzed in DfM~\cite{wang2022dfm}, BEVDet4D~\cite{Bevdet4d}, MV-FCOS3D++~\cite{wang2022mvfcos3d++} and recent follow-ups~\cite{wang2022sts,li2022bevstereo,solofusion,videobev,dort,badet}.
\end{itemize}

\begin{table}[btp]
\footnotesize
  \centering
  \caption{Results of depth-based PV to BEV methods on the KITTI 3D object detection benchmark.}
  \vspace{-2ex}
  \resizebox{0.75\linewidth}{!}{ 
    \begin{tabular}{c|c|c|c}
    \hline
    \multicolumn{1}{c|}{Methods} & \multicolumn{3}{c}{KITTI Performance(\%)} \\
\cline{2-4}    \multicolumn{1}{c|}{} & Easy  & Moderate  & Hard \\
    \hline
    PL(Mono)~\cite{pseudolidar} & 9.87  & 6.4   & 5.46 \\
           PatchNet~\cite{patchnet} & 15.68 & 11.12 & 10.17 \\
           AM3D~\cite{AM3D}  & 16.5  & 10.74 & 9.52 \\
           OFT~\cite{OFT}   & 2.5   & 3.28  & 2.27 \\
           CaDDN~\cite{CADDN} & 19.17 & 13.41 & 11.46 \\
           ImVoxelNet~\cite{imvoxelnet} & 17.15 & 10.97 & 9.15 \\
           DfM w/o pose & 22.84 & 16.82 & 14.65 \\
           BEVDet~\cite{Bevdet} &  -    &  -    &  - \\
           M2BEV~\cite{M2BEV} &  -    &  -    &  - \\
           BEVDet4D~\cite{Bevdet4d} &  -    &  -    &  - \\
           BEVerse~\cite{beverse} &  -    &  -    &  - \\
\cline{1-4}           PL(Stereo) & 54.5  & 34.1  & 28.3 \\
           PL++~\cite{pseudolidar++}  & 61.1  & 42.4  & 37 \\
           E2E PL~\cite{e2e-pseudolidar} & 64.8  & 43.9  & 38.1 \\
           DSGN~\cite{dsgn}  & 73.5  & 52.18 & 45.14 \\
           LIGA-Stereo~\cite{liga} & 81.39 & 64.66 & 57.22 \\
    \hline
    \end{tabular}%
    }
    \vspace{-2ex}
  \label{tab:3ddet-kitti}%
\end{table}%

\begin{table}[htbp]
\footnotesize
  \centering
  \caption{Results of depth-based methods on the nuScenes 3D object detection benchmark}
  \vspace{-2ex}
  \scalebox{0.9}{
    \begin{tabular}{c|c|c|c|c|c|c|c}
    \hline
    \multirow{2}[4]{*}{Methods} & \multicolumn{7}{c}{nuScenes Performance(\%)} \\
\cline{2-8}          & mAP   & mATE  & mASE  & mAOE  & mAVE  & mAAE  & NDS \\
    \hline
    BEVDet & \multicolumn{1}{r|}{0.422} & \multicolumn{1}{r|}{0.529} & \multicolumn{1}{r|}{0.236} & \multicolumn{1}{r|}{0.395} & \multicolumn{1}{r|}{0.979} & \multicolumn{1}{r|}{0.152} & \multicolumn{1}{r}{0.482} \\
    M2BEV & \multicolumn{1}{r|}{0.429} & \multicolumn{1}{r|}{0.583} & \multicolumn{1}{r|}{0.254} & \multicolumn{1}{r|}{0.376} & \multicolumn{1}{r|}{1.053} & \multicolumn{1}{r|}{0.19} & \multicolumn{1}{r}{0.474} \\
    BEVDet4D & \multicolumn{1}{r|}{0.426} & \multicolumn{1}{r|}{0.56} & \multicolumn{1}{r|}{0.254} & \multicolumn{1}{r|}{0.317} & \multicolumn{1}{r|}{0.289} & \multicolumn{1}{r|}{0.186} & \multicolumn{1}{r}{0.552} \\
    BEVerse & \multicolumn{1}{r|}{0.393} & \multicolumn{1}{r|}{0.541} & \multicolumn{1}{r|}{0.247} & \multicolumn{1}{r|}{0.394} & \multicolumn{1}{r|}{0.345} & \multicolumn{1}{r|}{0.129} & \multicolumn{1}{r}{0.531} \\
    \hline
   
    \end{tabular}}
    \vspace{-2ex}
  \label{tab:3ddet-nus}%
\end{table}%

%% file: MLP.tex
\vspace{-2ex}
\section{MLP based PV to BEV}
\label{sec:MLP}

\begin{figure}
    \centering
    \includegraphics[width=0.5\textwidth]{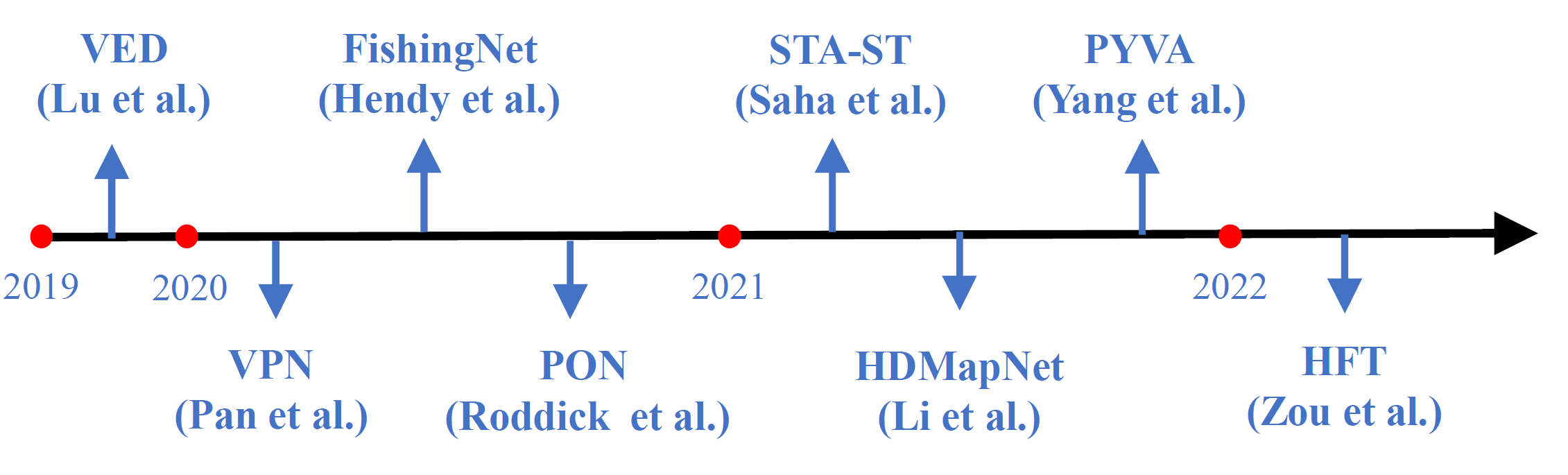}
    \vspace{-4ex}
    \caption{Chronological overview of MLP based PV to BEV methods.}
    \vspace{-2ex}
    \label{fig:MLP-overview}
\end{figure}

Multilayer Perceptron (MLP) is usually taken as a complex mapping function and has already made impressive achievement on mapping the input to the output with different modalities, dimensions, or representations. Escaping from inherit inductive biases contained in a calibrated camera setup, some methods~(Fig.~\ref{fig:MLP-overview}) tend to utilize the MLP to learn implicit representations of camera calibrations to transform between PV and BEV. 

\subsection{Basic Usage in Different Ways}
\begin{figure}[t]
\centering
\includegraphics[width=1\columnwidth]{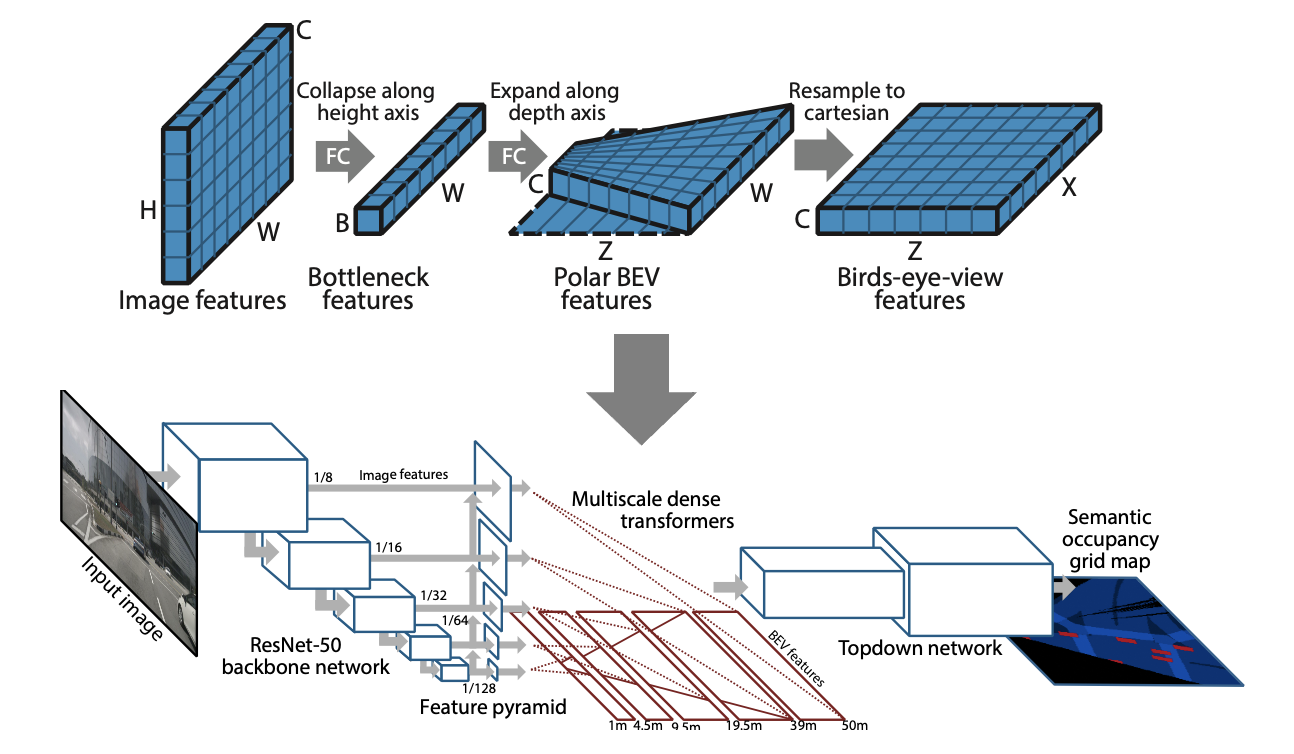}
\vspace{-4ex}
\caption{PON~\cite{PON} transforms PV features to BEV in a column-wise manner.}
\label{fig:PON}
\vspace{-2ex}
\end{figure}

VED~\cite{VED} employs a variational encoder-decoder architecture with 
an MLP bottleneck layer to transform the front-view visual information of the driving scene into the two-dimensional top-view Cartesian coordinate system. It is the first to perform end-to-end learning on monocular images to produce a semantic-metric occupancy grid map in real time. 
Motivated by the need for a global receptive field, VPN~\cite{VPN} chooses a two-layer MLP to transform each PV feature map to a BEV feature map through a flattening-mapping-reshaping process. It then adds all the feature maps from different cameras for multi-view fusion.
 Based on the view transformation module of VPN, FishingNet~\cite{FishingNet} converts the camera features to BEV space and conducts late fusion with radar and LiDAR data for multi-modal perception and prediction. 
To fully use the spatial context and better focus on small objects such as pedestrians, PON~\cite{PON} and STA-ST~\cite{STA-ST} first take advantage of a feature pyramid~\cite{FPN} to extract image features at multiple resolutions, as shown in Fig.~\ref{fig:PON}. Then the view transformation is performed by collapsing the image features along the height axis and expanding along the depth axis through MLP.  This design is inspired by the observation that while the network needs a lot of vertical context to map features to BEV~(due to occlusion, lack of depth information, and the unknown ground topology), in the horizontal direction, the relationship between BEV locations and image locations can be established using simple camera geometry. Such a column-wise view transformation idea is also explored in the transformer-based PV-to-BEV methods, as shown in Sec.~\ref{sec:transformer-based}.

Also adopting the MLP-based feature projection strategy, HDMapNet~\cite{Hdmapnet} aims to produce vectorized map elements in BEV and instance embedding and directions from images of the surrounding cameras. The unidirectional projection makes it difficult to guarantee the frontal-view information is delivered effectively, so an extra MLP can be used to project the feature from BEV back to PV to check whether it is correctly mapped. Motivated by this bidirectional projection, PYVA~\cite{PYVA} proposes a cycled self-supervision scheme to consolidate the view projection. It further introduces an attention-based feature selection process to correlate both views to get a stronger BEV feature for downstream segmentation tasks.

\subsection{Combination with Previous Streams}
HFT~\cite{HFT} gives an analysis of the pros and cons of camera model-based feature transformation and camera model-free feature transformation. The former, IPM-based methods, can easily handle PV-to-BEV transformation in regions such as local roads and carparks, but these methods rely on the flat-ground assumption, making distortions for those regions above the ground. The latter, MLP-based or attention-based methods, can avoid the basis, but they converge slowly without any geometric priors.  To benefit from both approaches and avoid their inherent drawbacks, HFT designs a hybrid feature transformation consisting of two branches to utilize the geometry information and capture global context respectively.

\begin{table}
\footnotesize
\caption{Results of MLP-based methods on BEV semantic segmentation task on the nuScenes val set~(\textbf{front-view-only}).}
\vspace{-3ex}
\begin{center}
\scalebox{0.85}{
    \begin{tabular}{c|c|c|c|c|c}
    \hline
    \multicolumn{1}{c|}{Methods} & \multicolumn{5}{c}{nuScenes Performance}\\
    \cline{2-6}
     ~ & Car  & Drivable & Crossing & Walkway & Carpark \\
    \hline
     
     VED~\cite{VED}  & 8.8  & 54.7   & 12.0 & 20.7 & 13.5 \\
     VPN~\cite{VPN}  & 25.5  & 58.0   & 27.3 & 29.4 & 12.3 \\
     PON~\cite{PON}  & 24.7  & 60.4   & 28.0 & 31.0 & 18.4 \\
     STA-ST~\cite{STA-ST} & 36.0  & 70.7  & 31.1 & 32.4 & 33.5\\
     HFT~\cite{HFT}  & 30.6  & 55.9   & 35.6 & 35.4 & 23.2 \\

    \hline
    \end{tabular}
    }
\end{center}
\label{tab: network-based-nus-seg-frontonly}
\vspace{-2ex}
\end{table}

\begin{table}
\scriptsize
\caption{Results of transformer-based PV to BEV methods on BEV semantic segmentation task on the nuScenes val set~(\textbf{surround view}). Noted that ``Drivable" is also called ``Road" in some papers and ``Lane" is marked with ``*" because different works might adopt different definitions of Lane.} 
\vspace{-3ex}
\begin{center}
\scalebox{0.9}{
    \begin{tabular}{c|c|c|c|c|c|c|c}
    \hline
    \multicolumn{1}{c|}{Methods} & \multicolumn{7}{c}{nuScenes Performance}\\
    \cline{2-8}
     & Car & Vehicle & Drivable & Lane* & Crossing & Walkway & Carpark \\
    \hline

 Image2Map  & 39.9 & 38.9 & 78.9 & - & - & - & - \\
    BEVFormer & 44.8 & 44.8 & 80.1 & 25.7 & - & - & - \\
    CVT  & - & 36.0 & 74.3 & - & - & - & - \\
    PETRv2  & - & 51.7 & 79.9 & 45.9 & - & - & -\\
    Ego3RT  & - & - & 79.6 & 47.5 & 48.3 & 52.0 & 50.3 \\
    GKT  & - & 38.0 & - & - &  - & - & - \\
    LaRa  & - & 35.4 & - & -  & - & - & - \\
    PolarFormer & - & - & 82.6 & 46.2 & 54.3 & 59.4 & 56.7 \\
    \hline
    \end{tabular}}
\end{center}
\label{tab: network-based-nus-seg-surround}
\vspace{-4ex}
\end{table}

\subsection{Summary}
MLP-based methods ignore the geometric priors of calibrated cameras and utilize MLP as a general mapping function to model the transformation from perspective view to bird’s eye view. Although MLP is theoretically  a universal approximator~\cite{Kim2003ApproximationBF}, the view transformation is still difficult to be reasoned due to the lack of depth information, occlusion, and so on. Moreover, the multi-view images are usually transformed individually and fused in a late-fusion manner, which prevents MLP-based methods from leveraging the geometric potential brought by the overlap regions. 
Table~\ref{tab: network-based-nus-seg-frontonly} shows the results achieved by MLP-based PV-to-BEV methods. We can observe that:
\begin{itemize}
	\item MLP-based methods pay more attention to the single-image case, while the multi-view fusion is still not fully explored.
	\item MLP-based methods are generally surpassed by re-cently proposed transformer-based methods, which will be described in the next section.
\end{itemize}


%% file: Transformer.tex
\begin{figure*}
    \centering
    \includegraphics[width=1.0\textwidth]{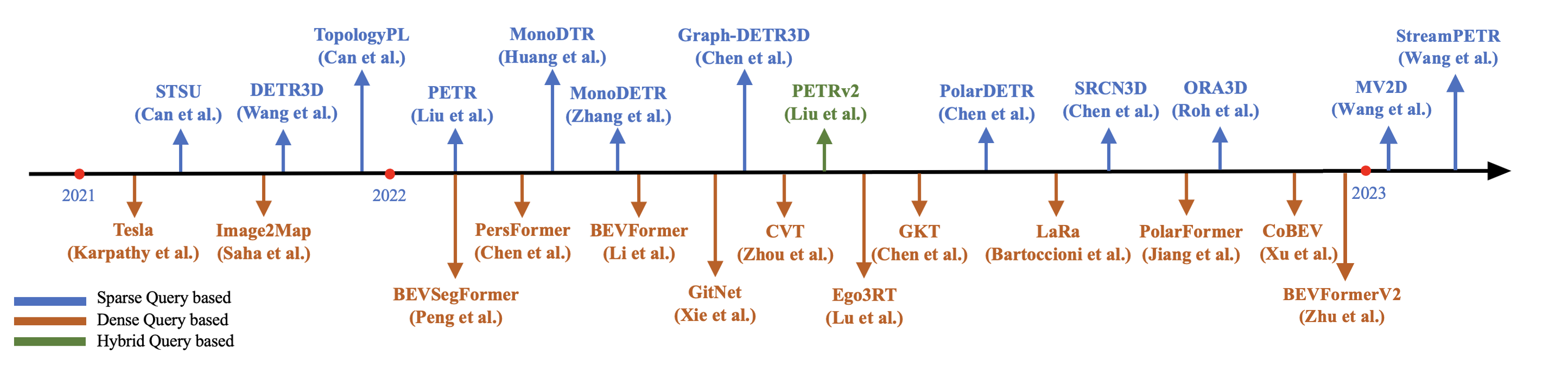}
    \vspace{-6ex}
    \caption{Chronological overview of transformer based PV to BEV methods.}
    \vspace{-2ex}
    \label{fig:transformer-based-overview}
\end{figure*}

\section{Transformer based PV to BEV}
\label{sec:transformer-based}

In addition to the aforementioned methods, transformer (with cross attention) is also a ready solution for mapping perspective view to bird’s eye view~(Fig.~\ref{fig:transformer-based-overview}). Although both employing the neural networks as the view projector for PV-to-BEV conversion without \textbf{explicitly} leveraging the camera model, there are three major differences between MLP-based and transformer-based tensor mapping. First, because the weighting matrix is fixed during inference, the mapping learned by MLP is not data dependent; in contrast, the cross attention in transformer is data dependent where the weighting matrix is dependent on the input data. This data dependency property makes transformer more expressive but hard to train. Second, the cross attention is permutation-invariant, meaning transformer needs positional encoding to distinguish the order of the input; the MLP is naturally sensitive to the permutation. Finally, instead of handling the view transformation in a forward way as done in MLP-based methods, transformer-based methods employ a top-down strategy by constructing queries and searching corresponding image features through an attention mechanism.

Tesla~\cite{tesla} is the first to project the perspective view features onto the BEV plane using transformers. This method first designs a set of BEV queries using positional encoding, then performs the view transformation through cross attention between BEV queries and the image features. Since then, many methods have been proposed to use transformers, or more specifically, the cross attention, for modeling the view transformation. Based on the granularity of learnable slots (called queries) in the transformer decoder, we divide the methods into three categories: sparse query-based, dense query-based and hybrid query-based. Next, we will introduce the representative works in each category and their pros and cons and then discuss the ways to involve geometric cues in these works.



\begin{figure}
	\centering
	    \subfigure[DETR]
	    {\centering
	     \begin{minipage}[t]{0.45\linewidth}
	     \includegraphics[width=1.0\columnwidth]{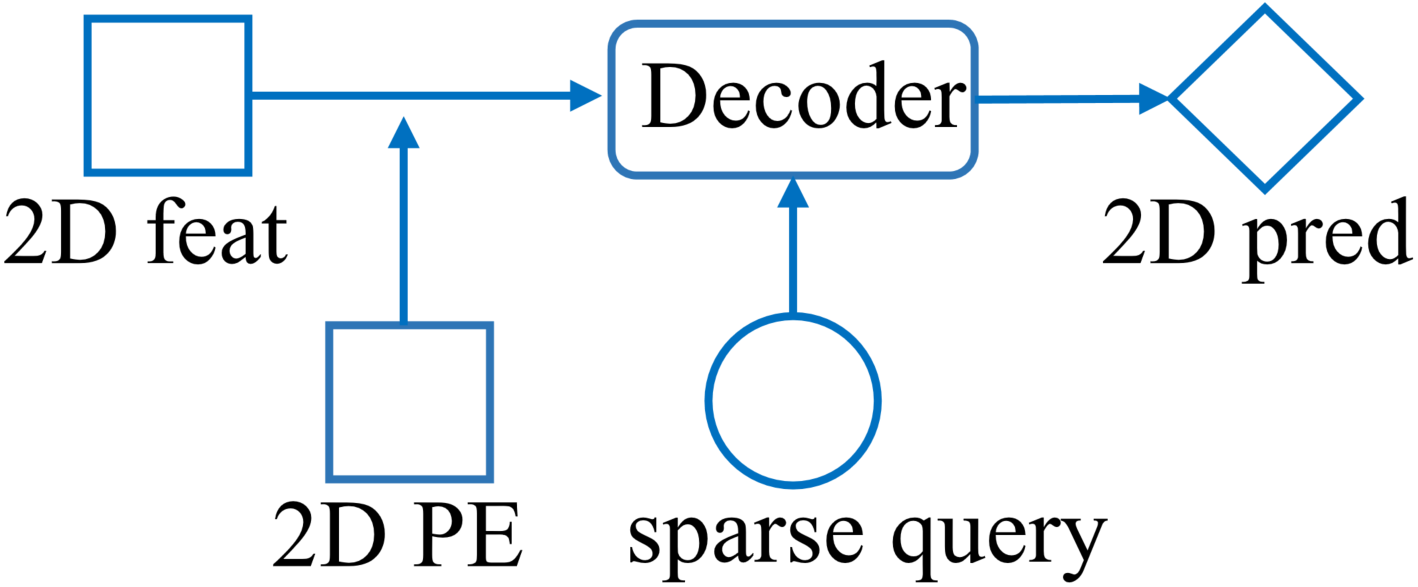} \end{minipage}}
	    \subfigure[DETR3D]
	    {\centering
	     \begin{minipage}[t]{0.45\linewidth}
	     \includegraphics[width=1.0\columnwidth]{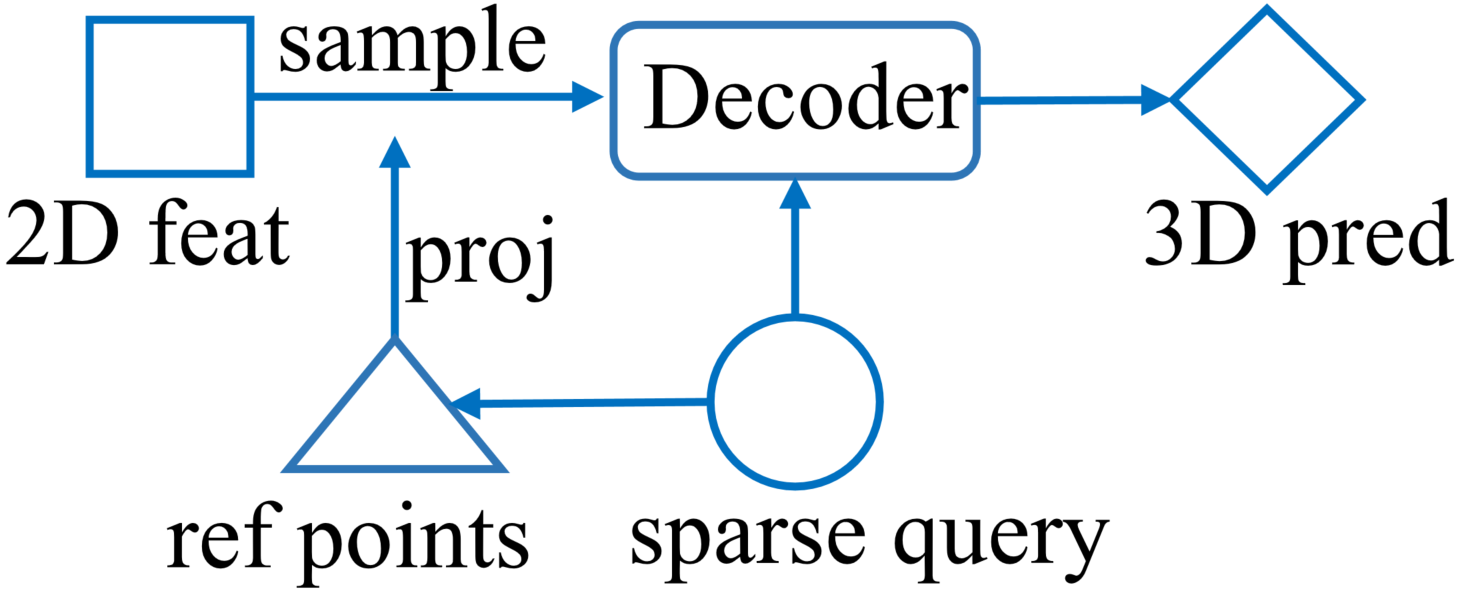} \end{minipage}}\\
	     \subfigure[PETR]
	    {\centering
	     \begin{minipage}[t]{0.45\linewidth}
	     \includegraphics[width=1.0\columnwidth]{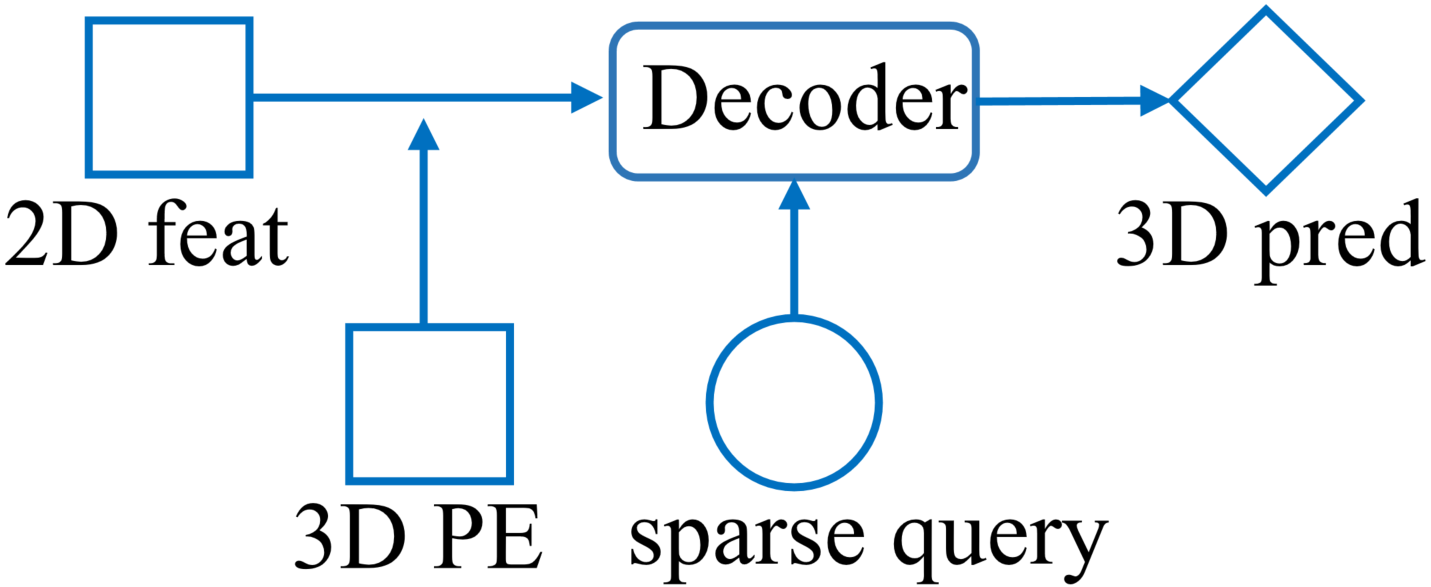} \end{minipage}}
	    \subfigure[dense query based]
	    {\centering
	     \begin{minipage}[t]{0.45\linewidth}
	     \includegraphics[width=1.0\columnwidth]{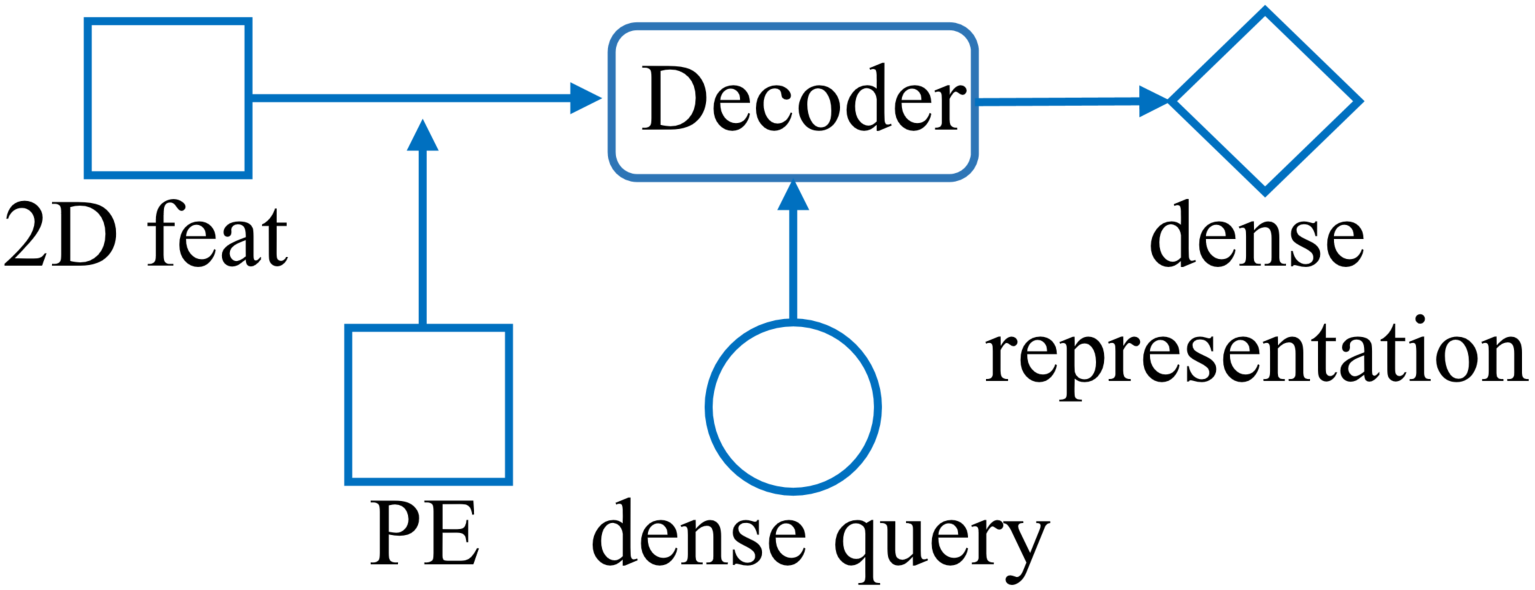} \end{minipage}}
	     \vspace{-2ex}
	     \caption{Paradigm comparison of DETR~\cite{detr}, DETR3D~\cite{detr3d}, PETR~\cite{petr}, and dense query-based methods.}
	     \vspace{-2ex}
	\label{fig:detr3d_vs_petr}
\end{figure}

\subsection{Sparse Query based Methods}

For sparse query-based methods, the query embeddings en-able the network to directly produce sparse perception results without explicitly performing the dense transformation of image features.
This design choice is natural for object-centric perception tasks such as 3D object detection but extending it towards dense perception tasks such as segmentation is not straightforward.

\begin{figure}[htbp]
\centering
    \centering
	\begin{minipage}[t]{0.85\linewidth}
	\includegraphics[width=1\columnwidth]{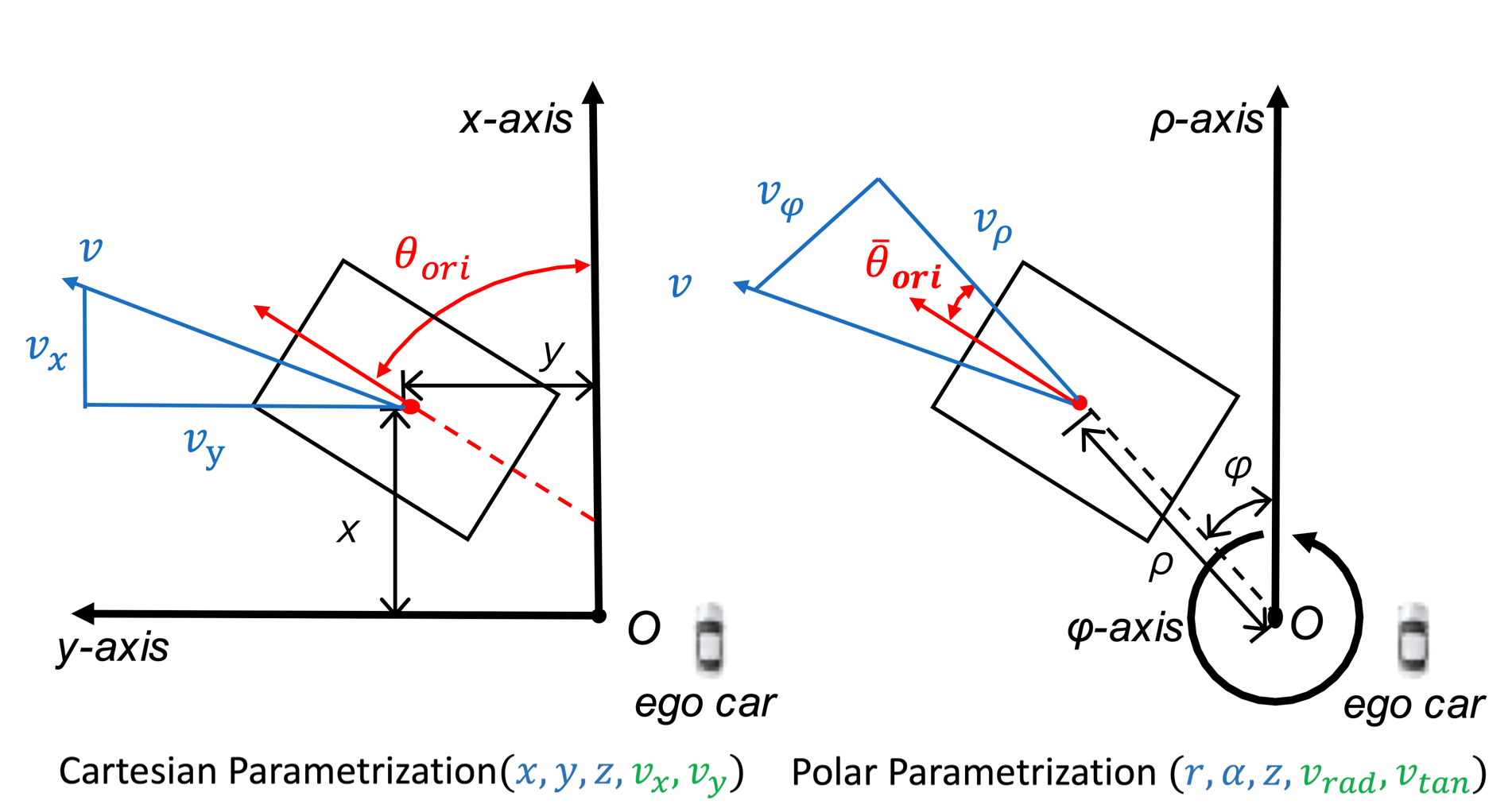} 
	\end{minipage}
\vspace{-2ex}
\caption{Illustration of parametrization of objects in PolarDETR~\cite{polardetr} and PolarFormer~\cite{polarformer}.}
\vspace{-2ex}
\label{fig:polar_coord}
\end{figure}

Inspired by the pioneering 2D detection framework DETR~\cite{detr}, STSU~\cite{stsu} follows the sparse query-based framework to extract the directed graph that represents the local road network in BEV space from a single image. This method can also detect 3D objects jointly by using two sets of sparse queries, one for centerline and one for dynamic object, where the dependency between objects and centerlines can be exploited by the network. The follow-up work TopologyPL~\cite{TopologyPL} improves STSU by considering the topology of the road network through preserving the minimal cycles.  
Concurrent with STSU, DETR3D~\cite{detr3d} proposes a similar paradigm but focuses on 3D detection for multi-camera input and replaces the cross attention by a geometry-based feature sampling process. It first predicts 3D reference points from the learnable sparse queries, then projects the reference points onto the image plane using the calibration matrices, and finally samples the corresponding multi-view multi-scale image features for end-to-end 3D bounding box prediction. Note that DETR3D relies on a geometric projection step similar with that in geometry-based PV-to-BEV methods in the previous sections, however, we categorize it into a transformer-based method since the key of their method is to leverage the transformer architecture to interact and enhance the sampled BEV features for better prediction. We provide a detailed discussion on how transformer-based methods leverage the geometric cues to inject the geometry relationship into their learning-based framework in Sec~\ref{subsec::geometric_cues}. To alleviate the complex feature sampling procedure in DETR3D, PETR~\cite{petr} encodes 3D positional embedding derived from camera parameters into 2D multi-view features so that the sparse queries can directly interact with the position-aware image features in vanilla cross attention, achieving a simpler and more elegant framework. A paradigm comparison of DETR3D and PETR is provided in Fig.~\ref{fig:detr3d_vs_petr}. The follow-up work PETRv2~\cite{petrv2} utilizes the temporal information by extending the 3D positional embedding to the temporal domain. 
To address the insufficient feature aggregation in DETR3D and improve the perception result in the overlap regions, Graph-DETR3D~\cite{graphdetr3d} enhances the object representation by aggregating various imagery information for each object query through graph structure learning. Similarly, ORA3D~\cite{ora3d} also focuses on improving the performance in the overlap regions of DETR3D. It regularizes the representation learning of overlap regions through stereo disparity supervision and adversarial training. 
To exploit the view symmetry of surround-view cameras as inductive bias to ease optimization and boost performance, PolarDETR~\cite{polardetr} proposes polar parameterization for 3D detection, which reformulates the bounding box parameterization, network prediction, and loss computation, all in the polar coordinate system, as shown in Fig.~\ref{fig:polar_coord}. It also leverages context features other than the features of projected reference points to alleviate the issue of insufficient contextual information in DETR3D.

SRCN3D~\cite{srcn3d}designs a sparse proposal-based multi-camera 3D detection method based on another 2D detection framework, SparseRCNN~\cite{sparsercnn}, where each proposal contains a learnable 3D bounding box and a learnable feature vector encoding the instance characteristics. To replace the cross attention-based feature interaction, a sparse feature sampling module and a dynamic instance interaction head are proposed to update the proposal features with the RoI features extracted from proposal boxes.

Instead of employing learnable object queries that remain fixed after training, MV2D~\cite{Wang2023ObjectAQ} presents a 2D-object-guided 3D detection framework that relies on sparse queries generated by a 2D object detector. By doing so, their approach can take advantage of all the enhancements made to off-the-shelf 2D detectors and can rely on sparser queries to recall the objects.

\subsection{Dense Query based Methods}

For dense query-based methods, each query is pre-allocated with a spatial location in 3D space or BEV space. The number of queries is decided by the spatial resolution of the rasterized space, which is usually larger than the number of queries in sparse query-based methods. The dense BEV representation can be achieved through the interaction between the dense queries and the image features for multiple downstream tasks such as 3D detection, segmentation, and motion prediction.

Tesla~\cite{tesla} first generates dense BEV queries in BEV space using positional encoding and context summary, then the view transformation is conducted with the cross attention between queries and multi-view image features. The vanilla cross attention between BEV queries and image features is performed without considering the camera parameters. To facilitate the geometric reasoning of cross attention, CVT~\cite{CVT} proposes a camera-aware cross-attention module that equips image features with positional embeddings derived from the cameras’ intrinsic and extrinsic calibration. Since the attention operation in each transformer decoder layer needs large memory complexity at vast query and key element numbers, the image resolution and BEV resolution are usually limited to reduce the memory consumption, which might hinder the model scalability in many cases. 

Recently, many efforts have been made to address this issue of dense query-based methods. Deformable attention~\cite{deformabledetr}, which combines the sparse spatial sampling of deformable convolution~\cite{deformableconv} and the relation modeling capability of attention~\cite{attention}, can remarkably reduce the memory consumption of the vanilla attention by only attending to sparse locations. 
It is adopted in the view transformation module of BEVSegFormer~\cite{bevsegformer} for BEV segmentation and of PersFormer~\cite{persformer} for 3D lane detection. Concurrently, BEVFormer~\cite{bevformer} also adopts the deformable attention for the interaction between dense queries located on the BEV plane and multi-view image features. It designs a set of history BEV queries and exploits the temporal cues through deformable attention between queries and history queries. The follow-up work BEVFormerV2~\cite{bevformerv2} introduces a perspective 3D detection head to adapt general 2D image backbones to the BEV model. Additionally, the proposals from the perspective head are fused with the original per-dataset object queries to make more accurate predictions. Ego3RT~\cite{ego3rt} places the dense queries on a polarized BEV grid and relies on deformable attention to make queries and multi-view image features interact. The polarized BEV features are then transformed into the Cartesian features through grid sampling,for downstream tasks. It is worth noting that instead of directly predicting reference points from query features in BEVSegFormer, BEVFormer and Ego3RT leverage the camera parameters and the pre-defined 3D positions of queries to compute the 2D reference points for feature sampling in the deformable attention. Similarly, PersFormer relies on IPM to compute the reference points on images. With such a design, the network could better identify the proper regions on the images with the geometric priors for guidance, but this risks them being more sensitive to the calibration matrices. 
GKT~\cite{gkt} unfolds kernel regions around the projected 2D reference points and interacts BEV queries with the corresponding unfolded kernel features, leading to a fixed mapping from BEV queries to pixel locations if the camera calibration is fixed. This operator can be regarded as a deformable attention with fixed sampling offsets and similarity-based attention weight. A BEV-to-2D look-up table indexing strategy is then proposed for fast inference. Instead of adopting deformable attention, CoBEVT~\cite{CoBEVT} proposes a novel attention variant called fused axial attention~(FAX), which reasons both high-level contextual information and regional detailed features with low computational complexity. Specifically, it first partitions the feature map into 3D non-overlapping windows, then performs local attention by attention within each local window and global attention by attention between different windows.

\begin{figure}[bp]
    \centering
    \vspace{-2ex}
	\includegraphics[width=0.85\columnwidth]{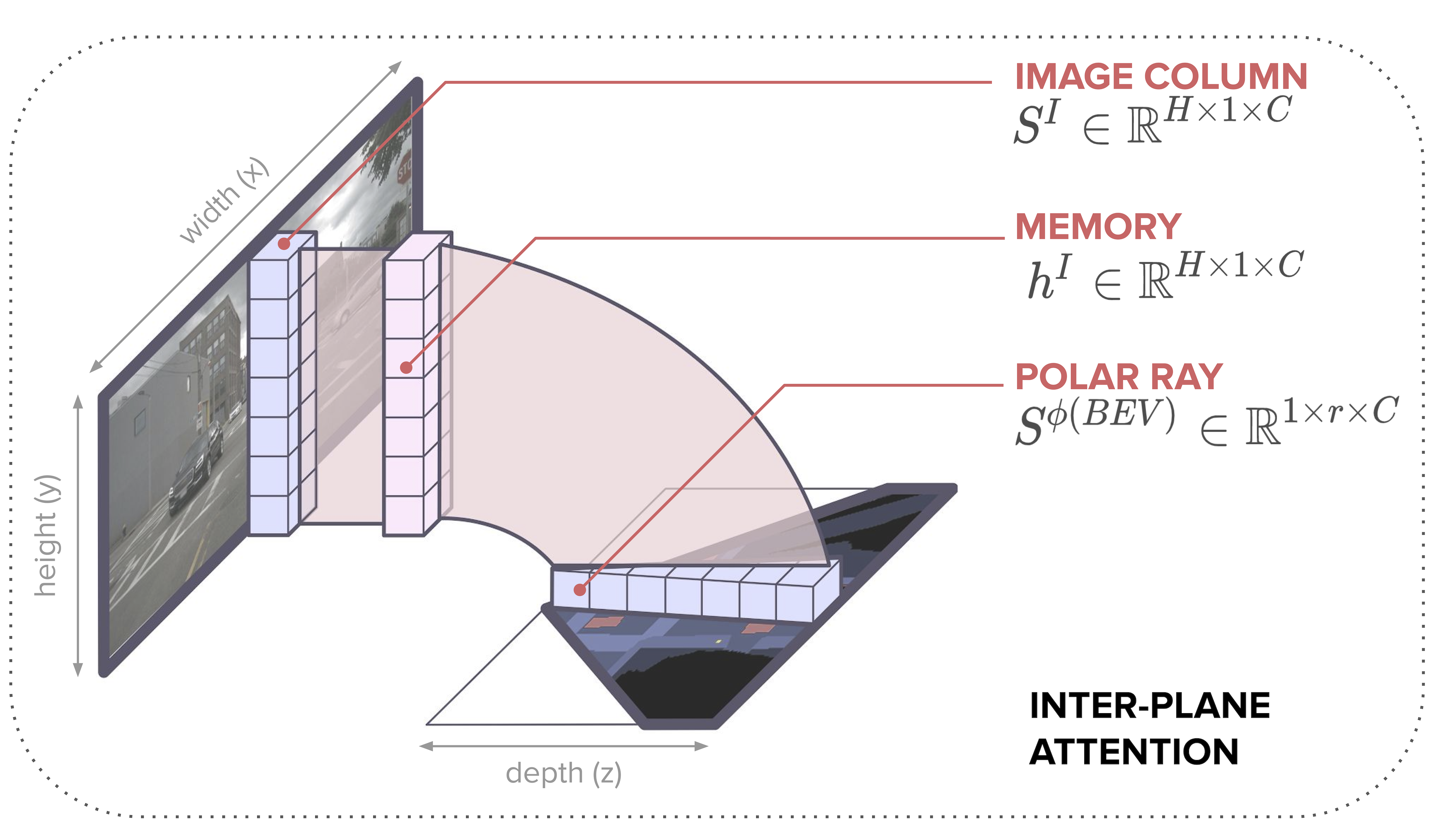} 
\caption{Image2Map: vertical scanlines in the image are passed separately to the transformer encoder to create a memory representation, which is decoded into a BEV polar ray.}
\vspace{-2ex}
\label{fig:image2map}
\end{figure}

Another promising way to reduce memory consumption is to simplify the cross attention-based interaction using 3D geometry constraints. Image2Map~\cite{image2map} proposes a framework for monocular BEV segmentation by first assuming the 1-1 relationship between vertical scanlines in the monocular image and the rays on the BEV plane starting from the camera center, as shown in Fig.~\ref{fig:image2map}. Then the view transformation can be formulated as a set of 1D sequence-to-sequence translation problems and modeled by a transformer. Based on this geometric constraint, Image2Map avoids the dense cross attention between 2D image feature maps and BEV queries, and instead only contains 1D sequence-to-sequence translation, leading to a memory-friendly and data-efficient architecture. This column-wise transformer module is adopted in GitNet~\cite{gitnet} as the second view transformation stage to enhance the initial BEV features obtained by the geometry-based first view transformation stage. Both Image2Map and GitNet only deal with a single image input. When considering 360-degree images as input, an additional adjustment is needed to align the polar rays from different cameras into the ego coordinate system, since the origins of polar rays are the different camera centers.
PolarFormer~\cite{polarformer} designs a polar alignment module to aggregate rays from multiple cameras to generate a structured polar feature map. A multi-scale polar representation learning strategy is proposed to handle unconstrained object scale variations over Polar's distance dimension. Similar to PolarDETR, the bounding box prediction is directly done in the polar coordinate system~(Fig.~\ref{fig:polar_coord}).

LaRa~\cite{lara} controls the computation footprint by first encoding the multi-view image features into a latent space and then obtaining the BEV features by querying the latent representation with a cross-attention module. The compact latent space is decoupled from the input size and output resolution, enabling precise computational budge control. In addition, a ray-based positional embedding derived from the calibration matrices is proposed to augment the visual features and guide the cross attention between features and the latent vectors.

\subsection{Hybrid Query based Methods}
Sparse query-based methods are suitable for object-centric tasks but cannot derive an explicit dense BEV representation, which makes them unsuitable for dense perception tasks such as BEV segmentation. Therefore, a hybrid query strategy is designed in PETRv2~\cite{petrv2}, where a dense segmentation query is proposed in addition to the sparse object query, and each segmentation query is responsible for segmenting a specific patch (i.e. of shape $16 \times 16$).

\subsection{Sparse query vs. Dense query}
Although sparse query-based methods achieve promising results on object detection tasks, their 3D representation has no sense of geometry structure w.r.t the ego coordinate frame, thus making it difficult for them to conduct dense prediction tasks such as map segmentation. In contrast, the dense query with explicit spatial distribution provides a dense and unified representation for the BEV space, which could be easily adopted by different perception heads. However, due to the huge computation cost under a large number of BEV queries, making the attention mechanism more efficient is necessary to achieve a high-resolution feature map. Efficient transformer architectures~\cite{Kitaev2020ReformerTE, Tay2020SparseSA, Wang2020LinformerSW} have gained intensive interest during the past few years. However, these works generally focus on self attention~\cite{monodtr}, where the key and query are derived from the same element set. Their effectiveness in cross attention, where the key and query come from two unaligned sets, remains under-explored.

\subsection{Combination with Previous Streams}
\label{subsec::geometric_cues}
Conceptually, the transformer-based PV-to-BEV methods can perform view transformation relying solely on the attention mechanism and do not necessarily need the geometric priors. Early methods~\cite{bevsegformer} indeed do not incorporate any geometric information, such as calibration matrices or per-pixel depth, into their transformer framework. However, the permutation-invariant nature makes transformer unaware of spatial relationships between image regions and BEV pixels, thus making the network slow to converge and data-hungry.
Thus more and more transformer-based methods are trying to involve 3D geometric constraints for fast convergence or data-efficiency. For example, the geometric projection relationship from Sec~\ref{sec::homograph} are commonly used in sparse-query-based transformer network for query feature sampling, and the depth supervision in Sec~\ref{sec::depth} are widely used in all kinds of transformer-based view projectors. 

\noindent\textbf{Geometric Projection / Calibration Matrices.} Given the 3D coordinates of the queries, the camera calibration matrices define the geometric projection from 3D space to image plane and vice versa, providing good cues for making the visual features and queries interact. Thus, the geometric projection relationship or the calibration matrices are leveraged in most of the transformer-based PV-to-BEV methods in various ways. Deformable attention-based methods~\cite{detr3d, graphdetr3d, bevformer, ego3rt} usually rely on the camera projection matrices to compute the 2D reference points for feature sampling, which helps the network attend to the proper regions on the images and get rid of global attention.
Another promising way to utilize the calibration matrices is to pre-assign each image vertical scanline to a BEV ray based on the camera geometry and then simplify the global cross attention into a column-wise attention, as is done in~\cite{image2map, gitnet, polarformer}. This strategy could also save computation significantly. 
In addition, calibration matrices can be used to generate 3D positional embedding~\cite{CVT, petr, petrv2, lara} to enrich the image feature with geometric priors and help the transformers to learn the mapping from perspective view to bird’s eye view with implicit geometric reasoning.

\noindent\textbf{Depth Information.} 
Although transformer-based PV-to-BEV methods do not necessarily need per-pixel depth for view transformation, the depth information is still shown to be important for geometric reasoning of transformers. 
On the nuScenes object detection benchmark, most transformer-based methods benefit from depth pre-training~\cite{FCOS3D, DD3D}, which provides useful depth-aware 2D features for establish-ing associations between queries and image features. 
Depth prediction can also be jointly optimized to assist the vision-centric 3D detection, where the ground truth depth can be derived from the projected LiDAR points~\cite{monodtr} or object-wise depth labels~\cite{monodetr}. Both MonoDTR~\cite{monodtr} and MonoDETR~\cite{monodetr} design a separate module to produce depth-aware features and predict per-pixel depth for positional encodings in transformers. MonoDTR then uses the transformer to integrate the context features and depth features for an anchor-based detection head, while MonoDETR modifies the transformer to be depth-aware to directly decode 3D predictions from 2D images by proposing a depth cross attention-equipped transformer decoder to make object queries interact with depth features.


\begin{table*}
\footnotesize
\caption{Results of transformer-based PV to BEV methods on the KITTI and nuScenes 3D object detection benchmark. "BEVFormer-S" and "PolarFormer-S" represent the model variants without temporal fusion.}
\vspace{-3ex}
\begin{center}
\scalebox{0.9}{
    \begin{tabular}{c|c|c|c|c|c|c|c|c|c|c|c|c}
    \hline
    \multicolumn{2}{c|}{\multirow{2}*{Methods}} & \multirow{2}*{Venue} & \multicolumn{3}{c|}{KITTI Performance (\%)} & \multicolumn{7}{c}{nuScenes Performance}\\
    \cline{4-13}
    \multicolumn{2}{c|}{} & ~ & Easy & Mod. & Hard & mAP & mATE & mASE & mAOE & mAVE & mAAE & NDS \\
    \hline
    \multirow{14}*{\tabincell{c}{Transformer\\Based\\ PV to BEV}} & DETR3D~\cite{detr3d} & CoRL 2021 & - & - & - & 0.412 & 0.641 & 0.255 & 0.394 & 0.845 & 0.133 & 0.479 \\
    ~ & PETR~\cite{petr} & ECCV 2022 & - & - & - & 0.441 & 0.593 & 0.249 & 0.383 & 0.808 & 0.132 & 0.504\\
    ~ & MonoDTR~\cite{monodtr} & CVPR 2022 & 21.99 & 15.39 & 12.73 & - & - & - & - & - & - & -\\
    ~ & MonoDETR~\cite{monodetr} & Arxiv 2022 & 25.00 & 16.47 & 13.58 & - & - & - & - & - & - & -\\
    ~ & PETRv2~\cite{petrv2} & Arxiv 2022 & - & - & - & 0.490 & 0.561 & 0.243 & 0.361 & 0.343 & 0.120 & 0.582\\
    ~ & Graph-DETR3D~\cite{graphdetr3d} & MM 2022 & - & - & - & 0.425 & 0.621 & 0.251 & 0.386 & 0.790 & 0.128 & 0.495\\
    ~ & PolarDETR~\cite{polardetr} & Arxiv 2022 & - & - & - & 0.431 & 0.588 & 0.253 & 0.408 & 0.845 & 0.129 & 0.493\\
    ~ & SCRN3D~\cite{srcn3d} & Arxiv 2022 & - & - & - & 0.347 & 0.723 & 0.278 & 0.472 & 0.986 & 0.158 & 0.412\\
    ~ & ORA3D~\cite{ora3d} & Arxiv 2022 & - & - & - & 0.423 & 0.595 & 0.254 & 0.392 & 0.851 & 0.128 & 0.489\\
    \cline{2-13}
    ~ & BEVFormer-S~\cite{bevformer} & ECCV 2022 & - & - & - & 0.435 & 0.589 & 0.254 & 0.402 & 0.842 & 0.131 & 0.495\\
    ~ & BEVFormer~\cite{bevformer} & ECCV 2022 & - & - & - & 0.481 & 0.582 & 0.256 & 0.375 & 0.378 & 0.126 & 0.569\\
    ~ & Ego3RT~\cite{ego3rt} & ECCV 2022 & - & - & - & 0.425 & 0.549 & 0.264 & 0.433 & 1.014 & 0.145 & 0.473\\
    ~ & PolarFormer-S~\cite{polarformer} & Arxiv 2022 & - & - & - & 0.455 & 0.592 & 0.258 & 0.389 & 0.870 & 0.132 & 0.503\\
    ~ & PolarFormer~\cite{polarformer} & Arxiv 2022 & - & - & - & 0.493 & 0.556 & 0.256 & 0.364 & 0.440 & 0.127 & 0.572\\
    \hline
    \end{tabular}}

\end{center}
\vspace{-4ex}
\label{tab: transformer-based-kitti-nus}
\end{table*}

\subsection{Summary}
Transformer-based view projectors are becoming more and more popular due to their impressive performance, strong relation modeling ability, and data-dependent property. In addition to being a view projector, transformer can also serve as a feature extractor to replace convolutional backbones or as a detection head to replace anchor-based, anchor-free heads. With the trend of developing big transformer models in NLP, researchers in the autonomous driving industry are also exploring the effectiveness of big and general transformers in extracting powerful representations for multiple tasks such as perception and prediction. On the other hand, the transformer-decoder-based detection head and the bipartite-matching-based label assignment strategy are 
now commonly adopted in image-based 3D detection, since this paradigm does not require post-processing such as NMS. 

nuScenes dataset is the most frequently used dataset for vision-centric perception with six calibrated cameras covering a 360-degree horizontal FOV. 
Table~\ref{tab: transformer-based-kitti-nus} and Table~\ref{tab: network-based-nus-seg-surround} show the results of transformer-based PV-to-BEV methods on detection and segmentation benchmark of nuScenes, respectively. Several observation can be drawn:
\begin{itemize}
	\item Dense queries are usually adopted when dense perception tasks (such as road segmentation) are considered, as the sparse query-based methods do not have an explicit representation of BEV space.
	\item As observed in depth-based view transformation methods, temporal information is also critical for transformer-based methods. Methods with temporal fusion~\cite{petrv2, bevformer, polarformer} generally outperform single-frame methods on mAP and mAVE by a large margin.
	\item As the perception range of each camera is a wedge with a radical axis, replacing perpendicular axis-based Cartesian parameterization with non-perpendicular axis-based polar parameterization~\cite{polardetr, polarformer}~(Fig.~\ref{fig:polar_coord}) is being proposed and would be an interesting direction for further investigation.
\end{itemize}

%% file: extension.tex
\section{Extension}
BEV representation of traffic scenarios, including precise localization and scale information, can accurately map to the real physical world, which facilitates many downstream tasks. Meanwhile, BEV representation also acts as a physical medium, providing an interpretable fusion way for the data from various sensors, timestamps, and agents. Furthermore, evolving from BEV perception, camera-based occupancy prediction with fine-grained semantic information has become a popular task and attract much attention now. In this section, we present three main extensions under BEV, multi-task learning strategies, fusion approaches, and semantic occupancy prediction. In particular, we also summarize empirical know-how to benefit future research works.

\label{extension}

\begin{table*}
\footnotesize
\caption{Results of joint learning of detection and segmentation on the nuScenes val set. Noted that multi-task version of Ego3RT is trained by finetuning the segmentation head with the pretrained detection model frozen.}
\vspace{-3ex}
\begin{center}
\scalebox{0.9}{
    \begin{tabular}{c|c|c|c|c|c|c|c|c|c|c|c}
    \hline
    \multirow{2}*{Methods} & \multicolumn{2}{c|}{Task Head} & \multicolumn{2}{c|}{3D Detection} & \multicolumn{7}{c}{BEV Segmentation~(IoU)}\\
    \cline{2-12}
    ~ & Det & Seg & NDS & mAP & Car & Vehicle & Drivable & Lane & Crossing & Walkway & Carpark \\
    \hline
    M$^2$BEV~\cite{M2BEV} & \checkmark & & 0.470 & 0.417 & - & - & - & - & - & - & - \\
    M$^2$BEV~\cite{M2BEV} &  & \checkmark & - & - & - & - & 77.2 & 40.5 & - & - & - \\
    M$^2$BEV~\cite{M2BEV} & \checkmark & \checkmark & 0.454 & 0.408 & - & - & 75.9 & 38.0 & - & - & - \\
    \hdashline
    BEVFormer~\cite{bevformer} & \checkmark & & 0.517 & 0.416 & - & - & - & - & - & - & - \\
    BEVFormer~\cite{bevformer} &  & \checkmark & - & - & 44.8 & 44.8 & 80.1 & 25.7 & - & - & - \\
    BEVFormer~\cite{bevformer} & \checkmark & \checkmark & 0.520 & 0.412 & 46.8 & 46.7 & 77.5 & 23.9 & - & - & - \\
    \hdashline
    PETRv2~\cite{petrv2} & \checkmark & & 0.496 & 0.401 & - & - & - & - & - & - & - \\
    PETRv2~\cite{petrv2} &  & \checkmark & - & - & - & 50.8 & 80.5 & 47.4 & - & - & - \\
    PETRv2~\cite{petrv2} & \checkmark & \checkmark & 0.495 & 0.401 & - & 49.4 & 79.1 & 44.3 & - & - & - \\
    \hdashline
    Ego3RT~\cite{ego3rt} &  & \checkmark & - & - & - & - & 79.6 & 47.5 & 48.3 & 52.0 & 50.3\\
    Ego3RT~\cite{ego3rt} & \checkmark & \checkmark & - & - & - & - & 74.6 & 36.6 & 33.0 & 42.6 & 44.1 \\
    \hdashline
    PolarFormer~\cite{polarformer} &  & \checkmark & - & - & - & - & 81.0 & 42.2 & 48.9 & 55.8 & 52.6\\
    PolarFormer~\cite{polarformer} & \checkmark & \checkmark & 0.465 & 0.388 & - & - & 82.6 & 44.5 & 50.1 & 57.4 & 54.1 \\
%
    \hline
    \end{tabular}
    }
\end{center}
\vspace{-4ex}
\label{tab:nus-multitask}
\end{table*}


\begin{figure}
	\centering
	    {\centering
	     \includegraphics[width=1.0\columnwidth]{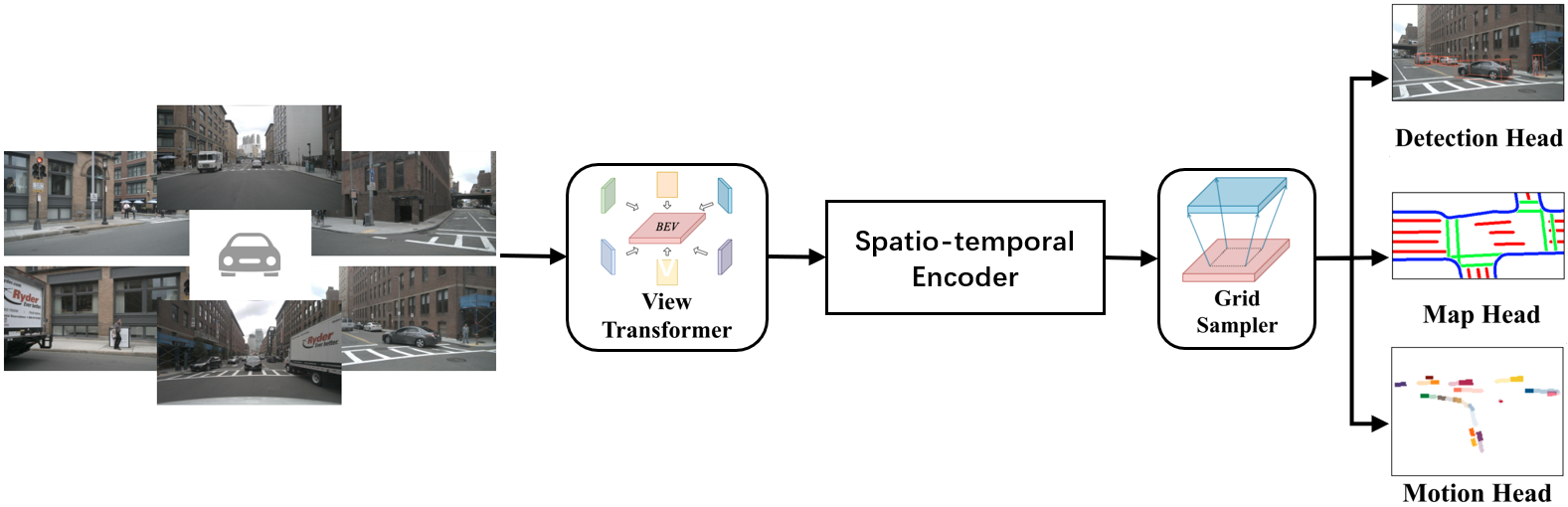}}
	 \vspace{-3ex}    
      \caption{BEVerse~\cite{beverse} uses a unified depth-based framework for multi-task learning from BEV.}
	     \vspace{-3ex}
	\label{fig:multi-task}
\end{figure}

\subsection{Multi-Task Learning under BEV}
The compact and effective BEV representation derived from PV2BEV methods is friendly to many downstream tasks, such as object detection, map segmentation, prediction, and motion planning. A shared backbone network can largely save computation cost and improve efficiency.  Thus several works attempt to use a unified framework to conduct multiple tasks simultaneously. 

With the assistance of spatio-temporal BEV representations from multi-camera videos, FIERY~\cite{FIERY} first proposes a frame-work for combining perception and prediction in one network. StretchBEV~\cite{StretchBEV} samples latent variables at each timestamp and estimates residual changes for producing future states. To reduce the memory consumption, BEVerse~\cite{beverse} designs iterative flow for efficient generation of future states and jointly reasons 3D detection, semantic map reconstruction, and motion prediction tasks Fig.~\ref{fig:multi-task}. M$^2$BEV~\cite{M2BEV} also proposes a multi-task approach based on BEV representation and simplifies the projection process by uniform depth assumption to save memory.
For transformer-based methods, STSU~\cite{stsu} and PETRv2~\cite{petrv2} introduce task-specific queries that interact with shared image features for different perception tasks.
BEVFormer~\cite{bevformer} first projects multi-view images onto the BEV plane through dense BEV queries and then adopts different task-specific heads such as Deformable DETR~\cite{deformabledetr} and mask decoder~\cite{panoticsegformer} over the shared BEV feature map for end-to-end 3D object detection and map segmentation. A similar strategy is also adopted in Ego3RT~\cite{ego3rt} and PolarFormer~\cite{polarformer}.

Although several works have shown that CNN benefits from joint optimizing with multiple related tasks, we observe that the joint training of 3D object detection and BEV segmentation usually does not bring improvement, as shown in Table~\ref{tab:nus-multitask}. The detection performance is usually hurt, and the improvement over segmentation performance is not consistent among different categories. More efforts are needed to explore the dependency between different perception tasks to achieve joint improvement.

\subsection{Fusion under BEV}
BEV representation provides a convenient way for multi-sensor, multi-frame, and multi-agent fusions, which greatly benefit perception in autonomous driving by utilizing comprehensive information. We summarize fusion methods relying on BEV representation in the following section according to different categories of source data.

 \begin{figure}[t]
\centering
\subfigure[LiDAR-Camera Fusion pipeline in UVTR.]
{\centering
 \includegraphics[width=0.92\columnwidth]{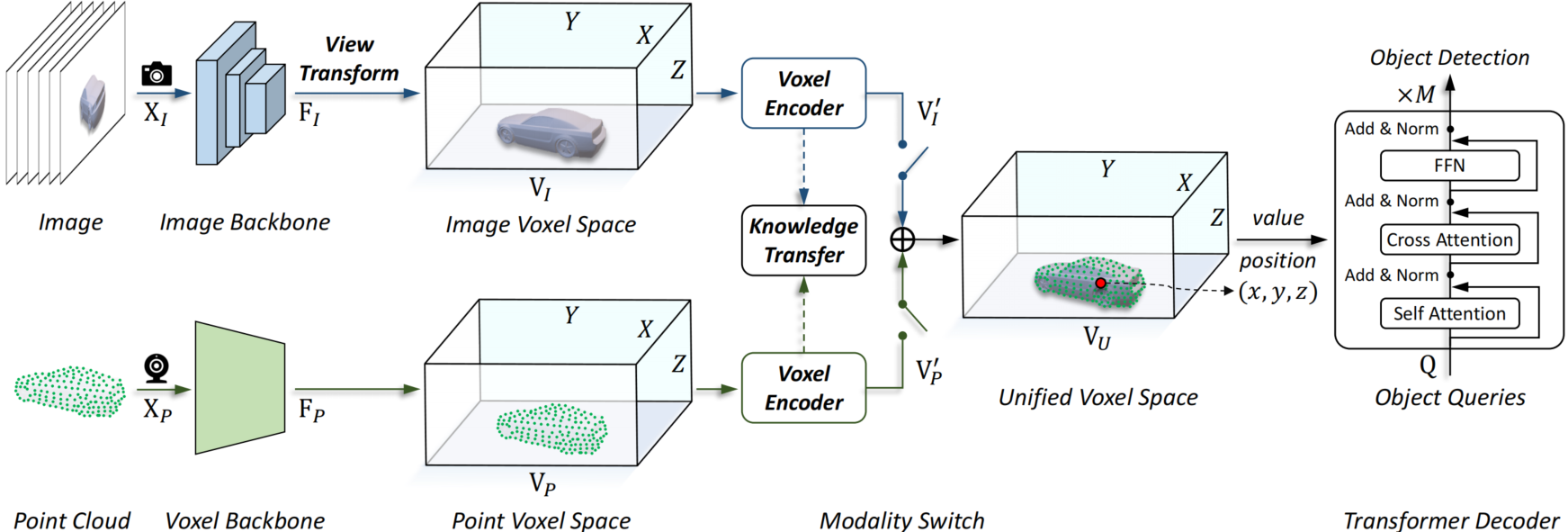}}\\
\subfigure[Multi-task Fusion framework in BEVFusion.]
{\centering
 \includegraphics[width=0.92\columnwidth]{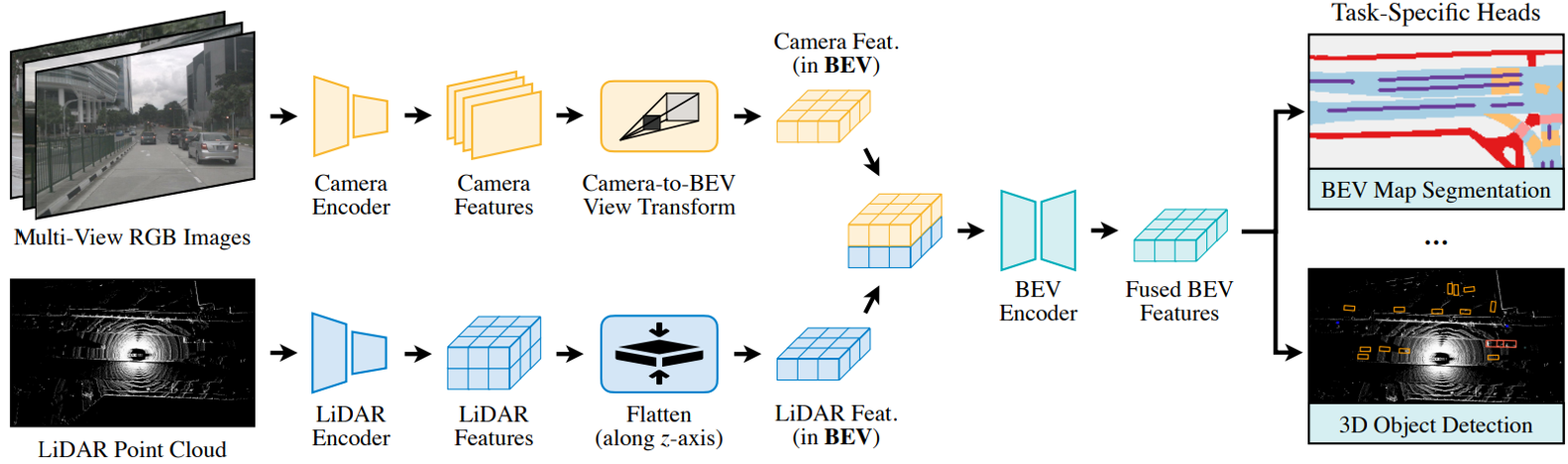} }\\
\subfigure[FUTR3D unified fusion framework.]
{\centering
 \includegraphics[width=0.92\columnwidth]{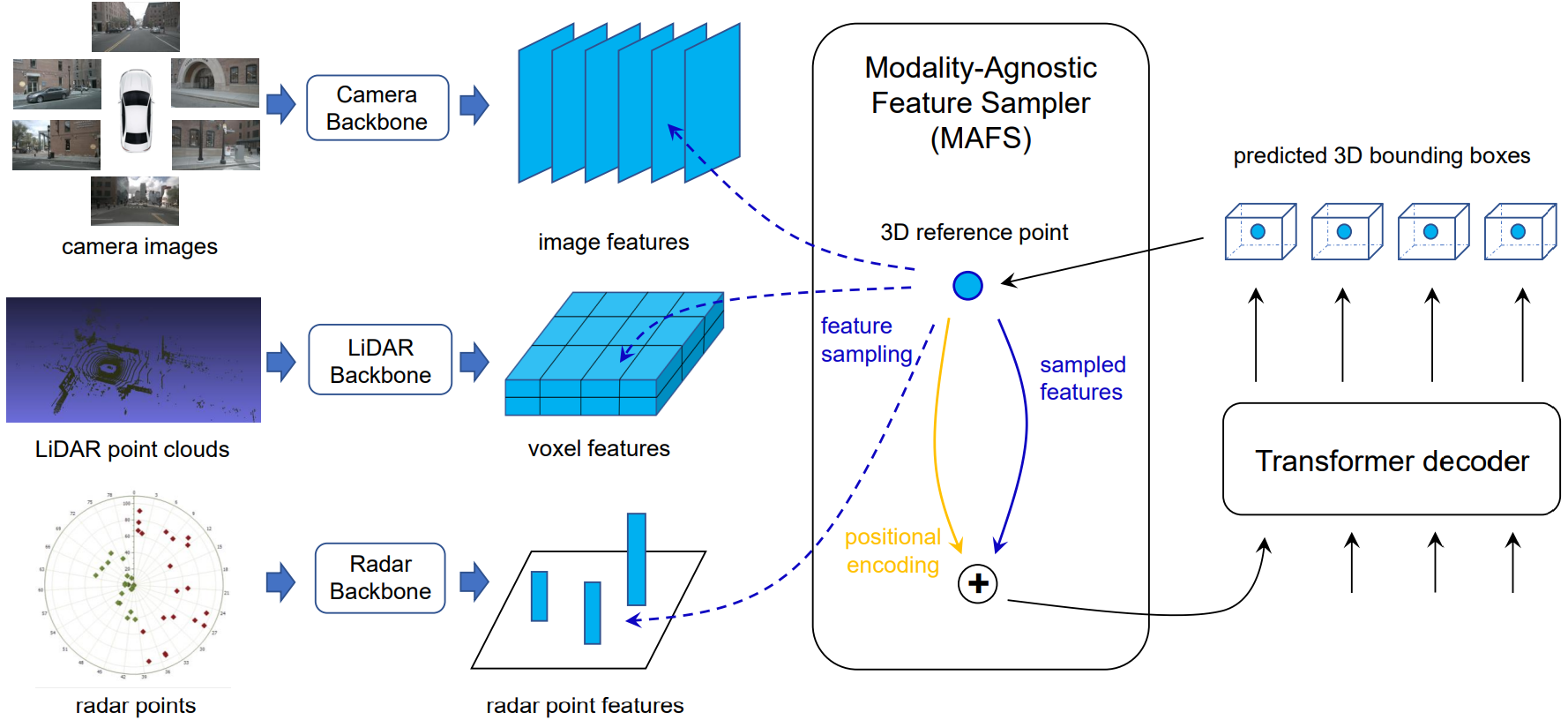}}
 \vspace{-2ex}  
\caption{LiDAR-image fusion pipeline on BEV space of UVTR~\cite{UVTR}, BEVFusion~\cite{BEVFusion}, and FUTR3D~\cite{FUTR3D}.}
\vspace{-4ex}
\label{fig:multi-modality}
\end{figure}

\subsubsection{Multi-Modality Fusion}

Current autonomous vehicles are usually equipped with three kinds of sensors, including Camera, LiDAR, and Radar, to conduct perception tasks. Different sensors have their own advantages and disadvantages, as shown in Table~\ref{tab:multi-modality}. Images captured by cameras are rich in appearance features such as colors, textures, and edges, but are sensitive to lighting and lack depth information. LiDAR point clouds contain accurate depth information and abundant geometry features, but they are short on texture information. Radar has a longer sensing range than LiDAR and can capture the velocity of moving objects directly, but the point clouds are extremely sparse and noisy, making it hard to extract shape and scale visual features. One ideal perception solution is integrating and utilizing all merits of these sensors in one network to achieve high-quality performance. However, due to the enormous differences in representations of raw data, reasonable and effective fusion is not easy.

Previous fusion strategies for images and point clouds can be classified into data-level fusion~\cite{PointPainting,PointAugmenting} and feature-level fusion~\cite{MV3D,4Dnet,Pointfusion,liang2018deep,ku2018joint}. The former uses the calibration matrix to attach pixel features to points and vice versa. The latter extracts image features in PV and point cloud features in 3D or BEV before directly fusing two kinds of high-dimensional features. With the fast development of BEV perception algorithms, a more inter-pretable way for image and point cloud fusion is transferring image features to BEV and fusing the features from two modal data according to the physical correspondences on BEV. 

Fusion methods can be further divided into three categories according to the exact fusion methods under BEV. The first class relies on the depth guidance and operates fusion in 3D space. UVTR~\cite{UVTR}, as shown in Fig.~\ref{fig:multi-modality} shows, constructs the voxel space by sampling features from the image plane according to predicted depth scores and geometric constraints. AutoAlign~\cite{AutoAlign} adaptively aligns semantic consistency between pixels and 3D voxels without explicit camera projections and guides cross-modal feature interactions through self-supervised learning. This is different from global-wise attention in AutoAlign. AutoAlignV2~\cite{AutoAlignV2} uses a deterministic projection matrix to guide the automatic alignment of cross-modal features and implements sparse sampling between modalities similar to ~\cite{deformabledetr}. Then, for each voxel, it is straightforward to establish the relationship between image features and associ-ated point cloud features. Also conducting fusion process in 3D space, Frustum PointNets~\cite{Frustumpointnets} and CenterFusion~\cite{CenterFusion} utilize frustum projection to transform image features of detected 2D objects to corresponding 3D locations and then fuse them with LiDAR detections and Radar detections, respectively. Methods in the second category perform a fusion operation on BEV features extracted from multi-modal inputs. BEVFusion~\cite{BEVFusion} fully retains the dense semantic information of the image and spatial geometry information during the fusion stage and proposes an efficient BEV pooling operation to speed up inference. RRF~\cite{RRF} defines a 3D volume for image features by projection and bilinear sampling, concatenates a rasterized Radar BEV image, and reduces the vertical dimension to finally get a BEV fused feature map. FISHINGNet~\cite{FishingNet} transforms features of cameras, LiDAR, and Radar into a single, common, and top-down semantic grid representation, respectively, and then aggregates these features for semantic grid predictions in BEV. The third kind of BEV fusion methods target 3D detection tasks by initializing 3D reference points as queries to extract features from all available modalities and conduct fusion operations. FUTR3D~\cite{FUTR3D} employs a query-based modality-agnostic feature sampler with a transformer decoder for sensor-fusion 3D object detection; this method can be easily adapted to any sensor combinations and setups. TransFusion~\cite{TransFusion} condenses the image features along the vertical dimension and then projects features onto the BEV plane using cross attention to fuse with the LiDAR BEV features. Specifically, such methods are output-oriented, and they learn where to fuse adaptively with the help of an attention mechanism. 
\begin{table*}[htbp]
  \centering
  \caption{Results of multi-modality BEV fusion methods on 3D detection task on the nuScenes val set. "L", "C" and "R" represent LiDAR, Camera, and Radar modality, respectively.}
  \vspace{-2ex}
  \scalebox{0.8}{
    \begin{tabular}{c|c|c|c|c|c|c|c|c|c|c|c|c|c|c|c}
    \hline
    \multicolumn{2}{c|}{\multirow{2}[4]{*}{Methods}} & \multirow{2}[4]{*}{Venue} & \multirow{2}[4]{*}{Modality} & \multicolumn{12}{c}{nuScenes Performance} \\
\cline{5-16}    \multicolumn{2}{c|}{} &       &       & \multicolumn{1}{c}{mAP} & \multicolumn{1}{c}{NDS} & \multicolumn{1}{c}{Car} & \multicolumn{1}{c}{Truck} & \multicolumn{1}{c}{C.V.} & \multicolumn{1}{c}{Bus} & \multicolumn{1}{c}{Trailer} & \multicolumn{1}{c}{Barrier} & \multicolumn{1}{c}{Motor.} & \multicolumn{1}{c}{Bike} & \multicolumn{1}{c}{Ped.} & T.C. \\
    \hline
    \multicolumn{2}{l|}{PointPainting~\cite{PointPainting}} & CVPR 2019 & LC    & 46.4  & 58.1  & 77.9  & 35.8  & 15.8  & 36.2  & 37.3  & 60.2  & 41.5  & 24.1  & 73.3  & 62.4 \\
    \multicolumn{2}{l|}{3D-CVF~\cite{PointPainting}} & ECCV 2020 & LC    & 52.7  & 62.3  & 83.0  & 45.0  & 15.9  & 48.8  & 49.6  & 65.9  & 51.2  & 30.4  & 74.2  & 62.9 \\
    \multicolumn{2}{l|}{FUTR3D~\cite{FUTR3D}} & Arxiv 2022 & LC    & 64.2  & 68    & 86.3  & 61.5  & 26    & 71.9  & 42.1  & 64.4  & 73.6  & 63.3  & 82.6  & 70.1 \\
    \multicolumn{2}{l|}{MVP~\cite{MVP}} & NeurIPS 2021 & LC    & 66.4  & 70.5  & 86.8  & 58.5  & 26.1  & 67.4  & 57.3  & 74.8  & 70    & 49.3  & 89.1  & 85 \\
    \multicolumn{2}{l|}{PointAugmenting~\cite{PointAugmenting}}  & CVPR 2021 & LC    & 66.8  & 71.0  & 87.5  & 57.3  & 28.0    & 65.2  & 60.7  & 72.6  & 74.3  & 50.9  & 87.9  & 83.6 \\
    \multicolumn{2}{l|}{FusionPainting~\cite{FusionPainting}} & ITSC 2021 & LC    & 68.1  & 71.6  & 87.1  & 60.8  & 30.0    & 68.5  & 61.7  & 7.8   & 74.7  & 53.5  & 88.3  & 85 \\
    \multicolumn{2}{l|}{UVTR~\cite{UVTR}} & Arxiv 2022 & LC    & 67.1  & 71.1  &  -    &  -    &  -    &  -    &  -    &  -    &  -    &  -    &  -    &  - \\
    \multicolumn{2}{l|}{TransFusion~\cite{TransFusion}} & CVPR 2022 & LC    & 68.9  & 71.7  & 87.1  & 60.0    & 33.1  & 68.3  & 60.8  & 78.1  & 73.6  & 52.9  & 88.4  & 86.7 \\
    \multicolumn{2}{l|}{BEVFusion~\cite{BEVFusion}} & Arxiv 2022 & LC    & 70.23  & 72.88  &  -    &  -    &  -    &  -    &  -    &  -    &  -    &  -    &  -    &  - \\
    \multicolumn{2}{l|}{AutoAlign~\cite{BEVFusion}} & IJCAI 2022 & LC    & 66.6  & 71.1  &  85.9    &  55.3    &  29.6    &  67.7    &  55.6    &  -    &  71.5    &  51.5    &  86.4    &  - \\   
    \multicolumn{2}{l|}{AutoAlignV2~\cite{AutoAlignV2}} & ECCV 2022 & LC    & 68.4  & 72.4  &  87.0    &  59.0    &  33.1    &  69.3    &  59.3    &  -    &  72.9    &  52.1    &  -    &  - \\
    \multicolumn{2}{l|}{CenterFusion~\cite{CenterFusion}} & WACV 2021 & RC   & 32.6  & 44.9  & 50.9  & 25.8  & -   & 23.4  & 23.5  & 48.4    & 31.4  & 20.1  & 37.0  & - \\
    \hline
    \end{tabular}%
    \vspace{-4ex}
  \label{tab:multi-modality}}%
\end{table*}%

\subsubsection{Temporal Fusion}
In addition to multi-modality fusion, temporal fusion is another critical component for a robust and reliable perception system for the following reasons. First, it accumulates sequential observations, which can alleviate the effect of self-occlusion and external occlusions caused by the view-dependent properties of camera. Second, temporal clues are necessity for estimating the temporal attributes of objects such as velocity, acceleration, steering, etc., which benefit the category classification and motion forecasting. Third, although depth estimation from a single image is naturally ill-posed and difficult, the stereo geometry formed by consecutive images provides important guidance and a well-studied theoretical basis for absolute depth estimation. 

Considering the benefits of temporal information existing in consecutive frames of images, many works~\cite{zhang2021temporal,wu2021towards,lin2019tsm} concatenate raw inputs, concatenate features extracted from images, or use RNN or transformer for video understanding, but they rarely take these steps for 3D perception. That is because cameras change poses as the ego-vehicle moves, meaning the consecutive perspective view representations do not have strict physical correspondence. Direct fusion for the temporal features in PV brings limited improvement to accurate 3D localization. Fortunately, BEV representation is easy to translate to the word coordinate system and can act as the bridge to fuse consecutive vision-centric data in a physical manner.


BEVDet4D~\cite{Bevdet4d} first warps the BEV feature map from previous frames into the current time based on ego-motion to put the features in the same coordinate sys-tem then concatenates the aligned feature maps along the channel dimension to feed into the detection head. Such concatenation-based temporal fusion strategies are simple and extendable and that is why they have also been adopted by other works such as Image2Map~\cite{image2map}, FIERY~\cite{FIERY}, and PolarFormer~\cite{polarformer}. In addition to concatenation, symmetric aggregation functions such as max, mean are also used to combine the temporally wrapped features~\cite{BEVstitch}. 
As moving objects can have different grid locations at different timestamps, BEV features from different times with the same physical positions might not belong to the same objects. Thus, to better build the association of the same objects from different times, BEVFormer~\cite{bevformer} models the temporal connection between features through a self-attention layer, where the current BEV features serve as query and the warped previous BEV features serve as key and value. 
In contrast, PETRv2~\cite{petrv2} performs the wrapping operation directly over the perspective view and 3D coordinate maps. It first generates the positional encoding of previous frames by converting the 3D coordinates of the previous frame into the current time based on ego-motion. Then the 2D image features and 3D coordinates of two frames are concatenated together for a transformer decoder, where the sparse object queries are able to interact with both current and previous features to get temporal information. Similarly, UniFormer~\cite{uniformer} converts PV features from previous frames into a unified virtual view and uses cross attention to fuse and integrate all the past and current features.
Instead of warping dense feature maps in previous methods, StreamPETR~\cite{Wang2023ExploringOT} propose an object-centric temporal mechanism which propagate the long-term historical information through a small number of object queries frame by frame, with only negligible storage and computation costs.
The aforementioned methods all focus on temporal fusion on a BEV plane, which happens after PV2BEV transformation. 
DfM~\cite{wang2022dfm}, instead, starts from a theoretical analysis in terms of the important role of temporal clues in depth estimation and chooses to exploit those clues in an earlier stage to facilitate the PV2BEV transformation through better depth estimation. Instead of relying on monocular understanding from a single image, DfM integrates the stereo geometric clues from temporally adjacent images.

As shown in Table~\ref{tab:nus-temporal}, lifting the models from the spatial-only 3D space to the spatial-temporal 4D space significantly improves the overall detection performance, especially for velocity and orientation prediction. However, most temporal models only leverage at most 4 previous frames, while the long-range history information is largely ignored. For example, the performance of BEVFormer begins to level off when the frame number is larger than 4, which means that the long-range information is not well exploited.

\begin{table*}
\footnotesize
\caption{Effect of temporal fusion on the nuScenes val set. }
\vspace{-4ex}
\begin{center}
\scalebox{0.9}{
    \begin{tabular}{c|c|c|c|c|c|c|c|c|c}
    \hline
    \multirow{2}*{Methods} & \multirow{2}*{Temporal} & \multirow{2}*{\#Frames} & \multicolumn{7}{c}{nuScenes Performance}\\
    \cline{4-10}
    ~ & ~ & ~ & mAP & mATE & mASE & mAOE & mAVE & mAAE & NDS \\
    \hline
    BEVDet~\cite{Bevdet} &  			& 1 & 0.312 & 0.691 & 0.272 & 0.523 & 0.909 & 0.247 & 0.392 \\
    BEVDet4D~\cite{Bevdet4d} & \checkmark & 2 & 0.323 & 0.674 & 0.272 & 0.503 & 0.429 & 0.208 & 0.453 \\
    \hdashline
    BEVFormer~\cite{bevformer} &  			& 1 & 0.375 & 0.725 & 0.272 & 0.391 & 0.802 & 0.200 & 0.448 \\
    BEVFormer~\cite{bevformer} & \checkmark & 4 & 0.416 & 0.673 & 0.274 & 0.372 & 0.394 & 0.198 & 0.517 \\
    \hdashline
    PETRv2~\cite{petrv2} &  			  & 1 & 0.384 & 0.775 & 0.270 & 0.470 & 0.605 & 0.189 & 0.461 \\
    PETRv2~\cite{petrv2} & \checkmark & 2 & 0.401 & 0.673 & 0.274 & 0.372 & 0.394 & 0.198 & 0.517 \\

    \hline
    \end{tabular}
    }
\end{center}
\vspace{-2ex}
\label{tab:nus-temporal}
\end{table*}

\subsubsection{Multi-agent Fusion}
Recent literature is mostly based on single-agent systems, which have trouble handling occlusions and detecting distant objects in complete traffic scenes. The development of Vehicle-to-Vehicle~(V2V) communication technologies makes it possible to overcome this issue by broadcasting the sensor data between nearby autonomous vehicles to provide multiple viewpoints of the same scene. Following this idea, CoBEVT~\cite{CoBEVT} first designs a multi-agent multi-camera perception framework that can cooperatively generate BEV map predictions. To fuse camera features from multi-agent data, it first geometrically warps the BEV features from other agents based on the ego pose and the pose of the sender, then fuses the information of received BEV features from multiple agents using a transformer with a proposed fused axial attention mechanism. However, since there is no available real-world dataset with multi-agent data, the proposed framework has only been validated on simulated datasets~\cite{carla}, and the real-world generalization capability is still unknown and needs further examination. 

\begin{table*}
\centering
\caption{Summary of implementation details of vision-centric BEV perception algorithms on nuScenes detection val set, including image resolution, image downsampling factor, BEV grid size, camera backbones, detection heads, auxiliary tasks, augmentation techniques, and class-balance sampling. GM, IDA, and BDA represent grid mask augmentation, image space data augmentation, and BEV space data augmentation, respectively. $\dagger$: the backbone is initialized from a FCOS3D backbone. }
\vspace{-2ex}
\tiny
\resizebox{\textwidth}{!}{
\begin{tabular}{l|c|c|c|c|c|c|c|c|c|c} 
\toprule
\textbf{Method}   & \textbf{Img-Resolution} & \textbf{Img-Stride} & \textbf{BEV-GridSize} &  \textbf{Backbone} & \textbf{Head} & \textbf{Auxiliary tasks} & \textbf{Augmentation} & \textbf{CBGS} & \textbf{mAP} & \textbf{NDS}\\
\midrule
FCOS3D~\cite{FCOS3D}  & 900$\times$ 1600 & 8,16,32,64,128  & -                     & R101-DCN          & FCOS3D          & -       & IDA          &             & 0.295 & 0.372 \\
DETR3D        & 900$\times$ 1600 & 8,16,32,64      & -                     & R101-DCN 		   & Deformable-DETR & -      & GM           &             & 0.303 & 0.374 \\
DETR3D        & 900$\times$ 1600 & 8,16,32,64      & -                     & R101-DCN$\dagger$ & Deformable-DETR & -     & GM           &  & 0.346 & 0.425 \\
DETR3D        & 900$\times$ 1600 & 8,16,32,64      & -                     & R101-DCN$\dagger$ & Deformable-DETR & -      & GM           & \checkmark  & 0.349 & 0.434 \\
BEVDet-R50    & 256$\times$ 704  & 16              & 0.8m$\times$ 0.8m     & R50      & CenterPoint     & -      & IDA, BDA     & \checkmark  & 0.298 & 0.379 \\
BEVDet-Tiny   & 256$\times$ 704  & 16              & 0.8m$\times$ 0.8m     & Swin-T   & CenterPoint     & -      & IDA, BDA     & \checkmark  & 0.312 & 0.392 \\
BEVDet-Base   & 640$\times$ 1600 & 16              & 0.4m$\times$ 0.4m     & Swin-B   & CenterPoint     & -      & IDA, BDA     & \checkmark  & 0.393 & 0.472 \\
BEVDet4D-Tiny & 256$\times$ 704  & 16              & 0.8m$\times$ 0.8m     & Swin-T   & CenterPoint     & -      & IDA, BDA     & \checkmark  & 0.323 & 0.453 \\
BEVDet4D-Base & 512$\times$ 1408 & 16              & 0.4m$\times$ 0.4m     & Swin-B   & CenterPoint     & -      & IDA, BDA     & \checkmark  & 0.396 & 0.515 \\
PETR-R101     & 384$\times$ 1056 & 16              & -                     & R101-DCN & DETR            & -      & GM, IDA, BDA & \checkmark  & 0.333 & 0.399 \\
PETR-R101     & 512$\times$ 1408 & 16              & -                     & R101-DCN & DETR            & -      & GM, IDA, BDA & \checkmark  & 0.357 & 0.421 \\
PETR-R101     & 512$\times$ 1408 & 16              & -                     & R101-DCN$\dagger$ & DETR            & -      & GM, IDA, BDA & \checkmark  & 0.366 & 0.441 \\
PETR-Tiny     & 512$\times$ 1408 & 16              & -                     & Swin-T   & DETR            & -      & GM, IDA, BDA & \checkmark  & 0.361 & 0.431 \\
BEVFormer-S   & 900$\times$ 1600 & 8,16,32,64      & 0.512m$\times$ 0.512m & R101-DCN$\dagger$ & Deformable-DETR & -      & GM           &             & 0.375 & 0.448 \\
BEVFormer     & 900$\times$ 1600 & 8,16,32,64      & 1.024m$\times$ 1.024m & R101-DCN$\dagger$ & Deformable-DETR & -      & GM           &             & 0.402 & 0.504 \\
BEVFormer     & 900$\times$ 1600 & 8,16,32,64      & 0.512m$\times$ 0.512m & R101-DCN$\dagger$ & Deformable-DETR & -      & GM           &             & 0.416 & 0.517 \\
BEVDepth-R50  & 256$\times$ 704  & 16              & 0.8m$\times$ 0.8m     & R50      & CenterPoint     & depth estimation      & IDA, BDA     & \checkmark  & 0.351 & 0.475 \\
BEVDepth-R101 & 512$\times$ 1408 & 16              & -                     & R101     & CenterPoint     & depth estimation      & IDA, BDA     & \checkmark  & 0.412 & 0.535 \\
\bottomrule
\end{tabular}}
\label{tab:implementation}
\vspace{-4ex}
\end{table*}

\subsection{Semantic Occupancy Prediction}
Vision-centric BEV perception has demonstrated an excellent trade-off between performance and efficiency. Based on it, the vision-centric semantic occupancy prediction task~\cite{tesla2022}, which aims at assigning semantic labels to every spatially-occupied region in 3D space, has gained increasing popularity in recent days due to the fine-grained geometry and semantic information in the occupancy representation.

We can categorize the relevant literature into two primary groups, \ie, voxel-sampling based methods and transformer-based methods. For the former type, it lifts the image representation to the voxel space through dense 2D-3D projections.
MonoScene~\cite{MonoScene} presents a 2D-to-3D feature sampling technique. It employs a series of stacked 2D/3D encoder-decoder blocks to capture the rich semantic information of the 3D space and designs a novel SSC loss to mitigate the occlusion ambiguity problem.
OccDepth~\cite{OccuDepth} recovers 3D spatial geometry by modeling dense correlations between stereo images. It proposes Stereo-SFA to enhance the feature correlation and integrates 3D geometry and semantic information through knowledge distillation.

As for transformer-based methods, these approaches explicitly construct a three-dimensional spatial representation and enhance the features by learning and searching for corresponding image features using transformers. TPVFormer~\cite{TPV} proposes a novel tri-perspective view (TPV), utilizing three mutually orthogonal planes to represent the 3D space. It projects 3D voxels onto the three views and fuses the associated features to model voxel characteristics. Similar to BEVFormer~\cite{bevformer}, TPVFormer creates three learnable grid planes, applies the cross-attention mechanism to learn image features, and employs cross-view attention to facilitate feature interaction among the three planes. Considering the sparsity of objects in 3D scenes, VoxFormer~\cite{VoxFormer} first estimates the depth of the image to obtain a set of occupied voxel queries and then utilizes the cross-attention mechanism to update the occupied voxels based on the image features. To alleviate the limitations imposed by the spatial invariance of 3D convolutions~\cite{MonoScene}, OccFormer~\cite{occformer} integrates global and local information along the horizontal direction to encode the 3D occupancy volume and introduces per-voxel embeddings and per-query embeddings in the occupancy decoder to relieve the problems of class imbalance and sparsity.

\begin{figure}
    \centering
    \includegraphics[width=0.85\columnwidth]{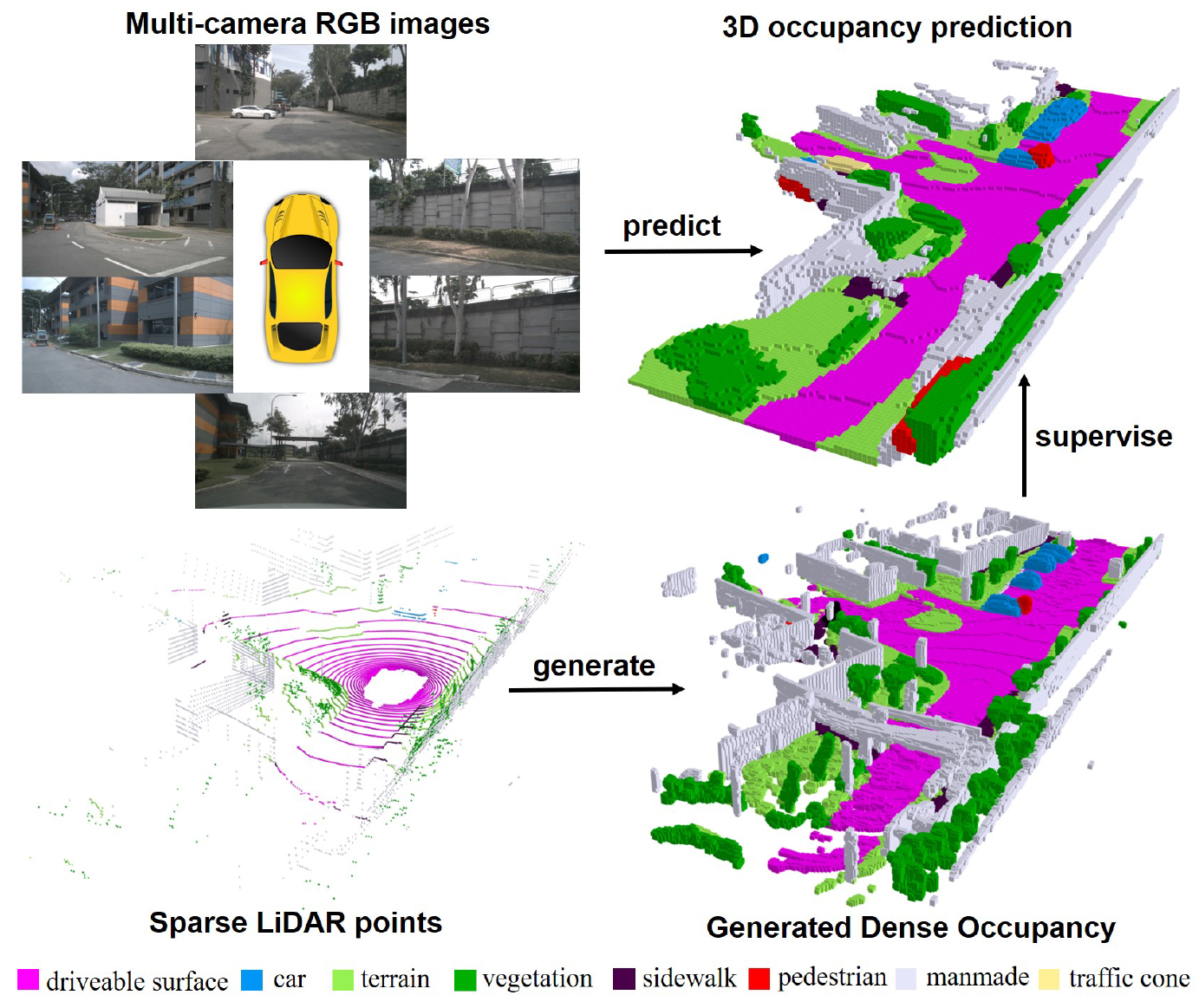}
    \caption{Overview of the semantic occupancy prediction task~\cite{SurroundOcc}. Based on the input of multi-camera images, methods need to predict volumetric occupancy of surrounding 3D scenes. The dense occupancy labels are usually generated by concatenating multiple LiDAR frames. }
    \vspace{-4ex}
    \label{fig:occupancy}
\end{figure}

Recently, several dense 3D occupancy prediction benchmarks have been established for boosting related research in the field of autonomous driving. Fig.~\ref{fig:occupancy} shows the overview of occupancy prediction task and the general way of label generation. OpenOccupancy~\cite{OpenOccupancy} uses the stacked LiDAR points of multiple frames to serve as semantic occupancy annotations. To handle the spatial sparsity and occlusion issues of LiDAR, it employs the AAP pipeline and pre-trained occupancy grid network to generate dense occupancy labels. 
SurroundOcc~\cite{SurroundOcc} extends the 2D-3D spatial attention mechanism to the multi-scale setting in order to improve the quality of 3D scene reconstruction. It takes multi-scale image features as input and employs a varying number of deformable cross-attention layers to extract the multi-scale 3D volume features. 
SCPNet~\cite{scpnet} introduces a novel completion network, that is built upon the designed Multi-Path Blocks (MPBs), to estimate both occupancy and semantic labels of the point cloud. It further presents the dense-to-sparse knowledge distillation strategy to fully use the rich temporal and semantic information in multiple frames.  

\subsection{Empirical Know-Hows}
\label{implementation}
This section presents empirical experiences of implementation details. Vision-centric perception methods usually involve multiple data modalities and conduct experiments on class-imbalanced datasets, thus requiring various data augmentation approaches with geometric relationship guaranteed and training tricks for categories with fewer annotations. In addition, balancing the trade-off of performance and efficiency is also an important problem. Next, we will discuss these details in four aspects: perception resolution, network designs, auxiliary tasks, and training details.

\subsubsection{Perception Resolution}

To perform view transformation from PV to BEV, the settings of the perception range for these two views are naturally critical to achieving an expected trade-off of performance and efficiency. The PV image resolution and the BEV grid size have increased significantly in recent years as the computational capability of graphics cards has made rapid progress. As shown in Table~\ref{tab:implementation}, the increase of these perception resolutions can significantly boost the performance, \emph{e.g.}, more 2\% mAP and NDS increase from $384\times1056$ to $512\times1408$ on PETR-R101. However, it also significantly affects the inference speed even with further optimization, \emph{e.g.}, 154.2ms to 37.9ms with PyTorch and 58.3ms to 18.4ms with TensorRT when reducing the input resolution from $640\times1760$ to $256\times704$ for BEVDet~\cite{bevpoolv2}. In particular, the inference time difference between $256\times704$ and $384\times1056$ versions is only \emph{3.4ms} with TensorRT-FP16, which means a practical and good new technique should achieve a similar performance improvement between these two versions (2\% NDS) with such a little computational overhead. Although these BEV-based methods achieve promising results on nuScenes, even approaching LiDAR-based methods, the high computational burden caused by high input resolutions is still a severe problem for deployment and is worthy of further exploration. In addition, here we mainly consider the influence of grid size on the BEV perception resolution because the perception range is always consistent with settings in LiDAR-based detectors~\cite{PointPillars,SECOND,CenterPoint}. However, these common settings are not enough in some practical scenarios, such as \emph{high-speed} cases on the expressway, which is another potential problem requiring future work.

\subsubsection{Network Designs}
Another critical factor in detection performance is the use of different feature extraction backbones and detection heads. As mentioned in recent works~\cite{detr3d,CADDN,wang2022mvfcos3d++,liga}, this type of method usually suffers from a lack of enough semantic supervision for perspective-view understanding. Therefore, most methods~\cite{detr3d,Bevdet,Bevdet4d,bevformer,petr} use PV backbones pretrained with monocular-based methods for 3D detection~\cite{FCOS3D,pgd} or depth estimation~\cite{DD3D}. \emph{Large backbone pretrained} with additional depth data or direct supervision from the perspective view can bring significant gains (more than 4\% mAP and NDS increase) for 3D detection performance, which is consistent with recent studies~\cite{pgd,monodle,arewemiss3dconf} of the crucial role of depth in this setting.
Furthermore, recent studies~\cite{geomim,hop,bevformerv2} further explore the potential of self-supervised learning and other 2D or 3D pre-training techniques tailored to geometry-related downstream tasks and shows promising performance.
As for the detection heads, apart from the conventional anchor-based 3D detection head, free-anchor head, and CenterPoint head used in LiDAR-based detection, transformer-based methods usually use a DETR3D or Deformable-DETR head to achieve a fully end-to-end design. Typically, \emph{anchor-based} 3D detection head is a more conventional and stable choice, while \emph{free-anchor head and CenterPoint head} usually achieve better performance in small objects~\cite{wang2022mvfcos3d++}. \emph{DETR-based} heads, in contrast, can naturally achieve sparse detection with more general formulations and thus attract more attention for academic exploration and large models.

\subsubsection{Auxiliary Tasks}
Due to various data modalities that can be leveraged during training such as images, videos, and LiDAR point clouds, the design of auxiliary tasks for better representation learning has also become a hot-spot issue in recent studies. In addition to classical auxiliary tasks like depth estimation~\cite{CADDN,liga}, monocular 2D and 3D detection~\cite{pgd,M2BEV}, and 2D lane detection~\cite{persformer}, several works also devise schemes for knowledge distillation from cross-modality settings such as monocular learn from stereo~\cite{pseudostereo} and stereo learn from LiDAR~\cite{SGM3D}.
BEVDepth~\cite{bevdepth} is an example in this direction, contributing a simple and \emph{economical} approach involving depth supervision in BEV methods with little computational overhead, resulting in 3\% NDS improvement on top of BEVDet~\cite{Bevdet}. However, such methods should be careful when tuning the \emph{loss weight} of auxiliary tasks, which is a sensitive hyper-parameter to make these techniques finally work.
In addition, this new trend still mainly focuses on experiments on small datasets, requiring further validation and development on \emph{large-scale} datasets where a large amount of training data may weaken the benefits of such training approaches.

\subsubsection{Training Details}
Finally, we would like to list several important details for tackling common issues in learning-based recognition. First, as most of these methods involve view transformation and different modalities, data augmentation can be applied to both perspective-view images and BEV grids. As shown in Table~\ref{tab:implementation}, recent methods usually exploit three types of data augmentations. Among them, \emph{BEV grid augmentation} is particularly important for this paradigm, which is also mentioned in \cite{Bevdet}. In addition, for \emph{class-imbalanced} issues, similar to LiDAR-based approaches, some methods~\cite{Bevdet,petr,bevdepth} exploit CBGS~\cite{CBGS} to increase the number of samples for long-tailed categories. It also empirically contributes to better convergence of such BEV models with more samples and longer training time. However, there are still very few works targeting this problem. More experiences from 2D and LiDAR-based perception are worthy of future work.

%% file: conclusion.tex
\section{Conclusion}
\label{conclusion}

This paper presents a comprehensive review of recent research on solving the view transformation between PV and BEV. We introduce and discuss related methods by clearly classifying them according to the core idea and downstream vision tasks. To facilitate further research and implementation, detailed comparison and analysis of performance and application scenarios are provided, and rich extensions of vision-centric BEV perception are also proposed.